\PassOptionsToPackage{unicode}{hyperref}
\PassOptionsToPackage{hyphens}{url}
\PassOptionsToPackage{dvipsnames,svgnames,x11names}{xcolor}
\documentclass[
  authoryear,
  preprint]{elsarticle}
\usepackage{xcolor}
\usepackage{amsmath,amssymb}
\usepackage{iftex}
\ifPDFTeX
  \usepackage[T1]{fontenc}
  \usepackage[utf8]{inputenc}
  \usepackage{textcomp} 
\else 
  \usepackage{unicode-math} 
  \defaultfontfeatures{Scale=MatchLowercase}
  \defaultfontfeatures[\rmfamily]{Ligatures=TeX,Scale=1}
\fi
\usepackage{lmodern}
\ifPDFTeX\else
\fi
\IfFileExists{upquote.sty}{\usepackage{upquote}}{}
\IfFileExists{microtype.sty}{
  \usepackage[]{microtype}
  \UseMicrotypeSet[protrusion]{basicmath} 
}{}
\makeatletter
\@ifundefined{KOMAClassName}{
  \IfFileExists{parskip.sty}{%
    \usepackage{parskip}
  }{
    \setlength{\parindent}{0pt}
    \setlength{\parskip}{6pt plus 2pt minus 1pt}}
}{
  \KOMAoptions{parskip=half}}
\makeatother

\makeatletter
\ifx\paragraph\undefined\else
  \let\oldparagraph\paragraph
  \renewcommand{\paragraph}{
    \@ifstar
      \xxxParagraphStar
      \xxxParagraphNoStar
  }
  \newcommand{\xxxParagraphStar}[1]{\oldparagraph*{#1}\mbox{}}
  \newcommand{\xxxParagraphNoStar}[1]{\oldparagraph{#1}\mbox{}}
\fi
\ifx\subparagraph\undefined\else
  \let\oldsubparagraph\subparagraph
  \renewcommand{\subparagraph}{
    \@ifstar
      \xxxSubParagraphStar
      \xxxSubParagraphNoStar
  }
  \newcommand{\xxxSubParagraphStar}[1]{\oldsubparagraph*{#1}\mbox{}}
  \newcommand{\xxxSubParagraphNoStar}[1]{\oldsubparagraph{#1}\mbox{}}
\fi
\makeatother

\usepackage{longtable,booktabs,array}
\usepackage{calc} 
\usepackage{etoolbox}
\makeatletter
\patchcmd\longtable{\par}{\if@noskipsec\mbox{}\fi\par}{}{}
\makeatother
\IfFileExists{footnotehyper.sty}{\usepackage{footnotehyper}}{\usepackage{footnote}}
\makesavenoteenv{longtable}
\usepackage{graphicx}
\makeatletter
\newsavebox\pandoc@box
\newcommand*\pandocbounded[1]{
  \sbox\pandoc@box{#1}
  \Gscale@div\@tempa{\textheight}{\dimexpr\ht\pandoc@box+\dp\pandoc@box\relax}
  \Gscale@div\@tempb{\linewidth}{\wd\pandoc@box}
  \ifdim\@tempb\p@<\@tempa\p@\let\@tempa\@tempb\fi
  \ifdim\@tempa\p@<\p@\scalebox{\@tempa}{\usebox\pandoc@box}
  \else\usebox{\pandoc@box}
  \fi%
}
\def\fps@figure{htbp}
\makeatother

\usepackage[]{natbib}
\usepackage{todonotes}

\usepackage{pdflscape}
\usepackage{algorithm}
\usepackage{algpseudocode}
\usepackage{booktabs}
\usepackage{longtable}
\usepackage{array}
\usepackage{multirow}
\usepackage{wrapfig}
\usepackage{float}
\usepackage{colortbl}
\usepackage{pdflscape}
\usepackage{tabu}
\usepackage{threeparttable}
\usepackage{threeparttablex}
\usepackage[normalem]{ulem}
\usepackage{makecell}
\usepackage{xcolor}
\makeatletter
\@ifpackageloaded{caption}{}{\usepackage{caption}}
\AtBeginDocument{
\ifdefined\contentsname
  \renewcommand*\contentsname{Table of contents}
\else
  \newcommand\contentsname{Table of contents}
\fi
\ifdefined\listfigurename
  \renewcommand*\listfigurename{List of Figures}
\else
  \newcommand\listfigurename{List of Figures}
\fi
\ifdefined\listtablename
  \renewcommand*\listtablename{List of Tables}
\else
  \newcommand\listtablename{List of Tables}
\fi
\ifdefined\figurename
  \renewcommand*\figurename{Figure}
\else
  \newcommand\figurename{Figure}
\fi
\ifdefined\tablename
  \renewcommand*\tablename{Table}
\else
  \newcommand\tablename{Table}
\fi
}
\@ifpackageloaded{float}{}{\usepackage{float}}
\floatstyle{ruled}
\@ifundefined{c@chapter}{\newfloat{codelisting}{h}{lop}}{\newfloat{codelisting}{h}{lop}[chapter]}
\floatname{codelisting}{Listing}

\makeatother
\makeatletter
\@ifpackageloaded{caption}{}{\usepackage{caption}}
\@ifpackageloaded{subcaption}{}{\usepackage{subcaption}}
\makeatother
\journal{Forensic Science International}
\usepackage{bookmark}
\IfFileExists{xurl.sty}{\usepackage{xurl}}{} 
\hypersetup{
  pdftitle={Normalisation-Based Likelihood Ratio Estimation for Forensic Authorship Verification},
  pdfauthor={Sadie Barlow; Andrea Nini; Edoardo Manino},
  pdfkeywords={Authorship Verification, Calibration, Likelihood Ratios},
  colorlinks=true,
  linkcolor={blue},
  filecolor={Maroon},
  citecolor={Blue},
  urlcolor={Blue},
  pdfcreator={LaTeX via pandoc}}

\begin{document}
\begin{frontmatter}
\title{Normalisation-Based Likelihood Ratio Estimation for Forensic
Authorship Verification}
\author[1]{Sadie Barlow%
\corref{cor1}
}
 \ead{sadie.barlow@manchester.ac.uk} 
\author[1]{Andrea Nini
}
 \ead{andrea.nini@manchester.ac.uk} 
\author[2]{Edoardo Manino
}
 \ead{edoardo.manino@manchester.ac.uk} 

\affiliation[1]{organization={The University of Manchester, Department
of Linguistics and English Language},addressline={Oxford
Road},city={Manchester},postcode={M13 9PL},postcodesep={}}
\affiliation[2]{organization={The University of Manchester, Department
of Computer Science},addressline={Oxford
Road},city={Manchester},postcode={M13 9PL},postcodesep={}}

\cortext[cor1]{Corresponding author}

\tnotetext[t1]{This work was supported by the North West Social Science Doctoral Training Partnership (NWSSDTP) [ES/P000665/1].}

\begin{abstract}
Authorship verification (AV) is the task of determining whether two texts
were written by the same author. In a forensic context, the strength of AV evidence can be quantified using
likelihood ratios. Most AV methods are score-based and deriving well-calibrated likelihood ratios
from these scores requires a separate calibration model. This, in turn, requires additional
amounts of case-relevant data, which is often time-consuming to obtain
and prepare. This study proposes two novel normalisation techniques, the
Square Root Correction and the Hapax Correction, for deriving likelihood
ratios from the AV method \textit{LambdaG} without the need of a calibration model
\citep{niniGrammarBehavioralBiometric2026}. These corrections are designed to mitigate the overestimation of evidential strength that may result from long or highly repetitive texts. Performance is evaluated
against logistic regression calibration across fifteen corpora and a
range of text lengths (100-9,500 tokens), using the log-likelihood ratio
cost (\(C_{llr}\)). The proposed methods achieve performance comparable
to logistic regression calibration, with the Hapax Correction
outperforming it in approximately 45\% of tests (weighted by corpora).
Furthermore, performance was more frequently close (within 5\%) when the
Hapax Correction was outperformed by logistic regression calibration,
compared with the reverse comparison. Eliminating the need to train a
calibration model reduces data-requirements, time and complexity,
thereby increasing the accessibility and transparency of forensic text
comparison. This combination of empirical performance and practical
advantages supports the adoption of the proposed methods in forensic
settings.
\end{abstract}


\end{frontmatter}

\section{Introduction}\label{introduction}

Across forensic science, the Likelihood Ratio Framework has emerged as
the 'logically and legally' endorsed standard for evaluation evidence
\citetext{\citealp[p.183]{ishiharaEstimatingStrengthAuthorship2022}; \citealp[p.26]{forensicscienceregulatorCodesPracticeConduct2021}}. The framework requires the comparison of the probability of evidence under two competing hypotheses. While the Likelihood Ratio Framework is firmly established in DNA analysis
\citep{baldingDNAProfileMatch1994a, taylorInterpretationSingleSource2013a}
and well-validated in forensic voice comparison
\citep{roseTechnicalForensicSpeaker2006a, morrisonMeasuringValidityReliability2011a},
its application in forensic text comparison remains comparatively
nascent
\citep{ishiharaStrengthForensicText2017, grantIdeaProgressForensic2022a, niniGrammarBehavioralBiometric2026}.

For score-based systems, Log-Likelihood Ratios (LLRs) can be derived through a process
called \textit{calibration}
\citetext{\citealp[p.174]{morrisonTutorialLogisticregressionCalibration2013}; \citealp{vandervloedInterchangeabilityCalibrationAudio2024}}. This involves training a separate calibration model, which learns the necessary transformation to align predicted probabilities with empirical outcome frequencies \citetext{\citealp{guoCalibrationModernNeural2017a}; \citealp[p.3215]{silvafilhoClassifierCalibrationSurvey2023a}; \citealp[p.2]{pullBayesianApproachProbability2025}}. 

In practice, calibration is most often achieved using logistic
regression
\citep{brummerApplicationindependentEvaluationSpeaker2006, gonzalez-rodriguezEmulatingDNARigorous2007, morrisonTutorialLogisticregressionCalibration2013}
(detailed in Section \ref{sec:cal}), which requires comprehensive and
case-specific training data. Compiling such datasets can often be complex
and computationally demanding. Data should ideally reflect the case
conditions \citep{morrisonBiGaussianizedCalibrationLikelihood2024a},
however, what this means is not obvious. 

To demonstrate whether the calibration data was sufficient to produce properly calibrated LLRs, additional tests must be performed. If a LLR is not properly calibrated, it is not considered to be valid and reliable forensic evidence \citetext{\citealp[p.164]{ramosReliableSupportMeasuring2013}; \citealp[p.14]{vergeerMeasuringCalibrationLikelihoodratio2021}}. In many forensic disciplines, including forensic text comparison, calibration is therefore an essential step for producing meaningful LLRs. 

Many cases of forensic text comparison concern authorship verification (AV). This is the task of determining whether a text of disputed authorship was
authored by the same individual as a text of known authorship
\citetext{\citealp[p.284]{koppelFundamentalProblemAuthorship2012a}; \citealp{pothaProfileBasedMethodAuthorship2014}; \citealp{juolaVerifyingAuthorshipForensic2021b}}.
AV is possible because individuals consistently reuse linguistic
patterns
\citep{coulthardAuthorIdentificationIdiolect2004, mollinEntirelyUnderstandBlairism2009, wrightUsingWordNgrams2017a},
and the unique combination of these patterns can be used to identify an
author.

A recently proposed AV method, LambdaG, is the focus of this study
\citep{niniGrammarBehavioralBiometric2026}. LambdaG is outlined in
Section \ref{sec:av}. This method returns a score in the form of an uncalibrated LLR.
To ensure that the score is interpretable, a separate post-hoc calibration is still required.

This paper investigates the nature of AV scores and examines whether
normalisation, the scaling of values to a common scale, offers a
functionally equivalent alternative to calibration in this context. In
Section \ref{sec:propm}, we introduce two novel normalisation techniques
- the Square Root Correction, which addresses the effect of
text length on AV scores, and the Hapax Correction, which adjusts scores
based on a measure of uniqueness relative to size.

These techniques are evaluated using 15 diverse datasets, spanning academic papers,
newspaper articles, blog posts, four corpora of reviews, Wikipedia talk
pages, two forum corpora, two email corpora, chat logs, text messages
, and tweets. For a subset of the corpora, the length of the text was systematically
varied, ranging from \(100\) to \(9,500\) tokens. This allowed us to test whether text length introduces a bias into the normalisation of scores. 

The results of this evaluation are summarised in Section \ref{sec:res}.
They indicate that, on average across the corpora, the best performing
normalisation technique (the Hapax correction) outperforms the baseline
of logistic regression calibration in 45.4\% of tests. In 56.91\% of
losses, this correction performed comparably to the baseline, remaining
within 5\% of logistic regression calibration.

By eliminating the need for a separate calibration model, whilst maintaining comparable performance, the limitations that calibration models entail are also removed. In particular, data requirements are reduced and in turn, the time cost and complexity of the method are lowered. In sum, this research contributes to the development of more accessible, efficient and interpretable methods for forensic AV.   

\section{Likelihood Ratios}\label{likelihood-ratios}

\label{sec:LLR}

A LLR is a measure of evidential value, quantifying the relative
probability of observing given evidence under two opposing hypotheses
\citep{biedermannReframingDebateQuestion2016}. It is defined as the
logarithm of the ratio of the probability of the observed evidence (\(E\)) if the
prosecution hypothesis (\(H_p\)) is true, to the probability of the
evidence if the defence hypothesis (\(H_d\)) is true \citep[eq.
1]{ishiharaScorebasedLikelihoodRatios2021a}:

\begin{equation}\phantomsection\label{eq-LLR}{ 
LLR = \log\frac{P(E|H_p)}{P(E|H_d)}
}\end{equation}

In the context of forensic AV, the two opposing hypotheses may be:

\begin{quote}
\(\mathbf{H}_p\): the candidate author wrote the disputed document
\end{quote}

\begin{quote}
\(\mathbf{H}_d\): the candidate author did not write the disputed
document
\end{quote}

The magnitude of the LLR represents the strength of the evidence in
support of one hypothesis over the other.

In forensic text comparison, however, leading methods typically output
similarity scores rather than calibrated LLRs
\citep{evertUnderstandingExplainingDelta2017, grieveAttributingBixbyLetter2019a}.
Currently, no AV method produces a well-calibrated LLR without a separate
calibration model. 

For example, the Impostors Method, which is one of the most widely known AV methods, produces a score between 0 and 1 that needs to be calibrated into a LLR \citep{koppelDeterminingIfTwo2014a}. Similarly, LUAR and STAR, state-of-the-art AV systems based on neural networks, produce cosine similarity scores, which, without calibration into a LLR, are suitable only for comparative evaluation \citep{rivera-sotoLearningUniversalAuthorship2021, huertas-tatoUnderstandingWritingStyle2024}. Furthermore, a recently proposed AV method, LambdaG, produces a score in the form of an uncalibrated LLR, a value that is constructed as a LLR but cannot be interpreted probabilistically without further processing \citep{niniGrammarBehavioralBiometric2026}. 

Therefore, all existing AV methods require calibration, as we explain in the following section. 

\section{Calibration}\label{calibration}

\label{sec:cal}

Calibration involves the adjustment of raw model scores through
transformations such as scaling and shifting
\citep[p.61]{morrisonEmpiricalEstimatePrecision2011}, so that the
resulting values allow meaningful LLR interpretation. For a perfectly
calibrated system, calculating the LLR of the systems output returns the
same value as the output itself
\citep{morrisonBiGaussianizedCalibrationLikelihood2024a}.

\subsection{Applied Calibration}\label{applied-calibration}

\label{sec:ac}

To elucidate the concept of calibration and its role in AV, consider the
analogy of weather forecasting
\citep{dawidWellCalibratedBayesian1982, degrootComparisonEvaluationForecasters1983}.
Intuitively, a well-calibrated weather forecasting system is one where
70\% of the time that there is a 70\% chance of rain, it rains
\citep[p.13]{degrootComparisonEvaluationForecasters1983}. The calibrated
forecast is therefore producing a prediction that corresponds to
observed relative frequencies
\citep[p.3215]{silvafilhoClassifierCalibrationSurvey2023a}, rendering it
suitable for probabilistic interpretation.

In contrast, if rain follows a 70\% forecast only 60\% of the time, the
model is systematically overestimating the likelihood of rainfall. In
this case, the reported probabilities cannot be interpreted as reliable
estimates of empirical frequency, indicating the need for calibration.

For LLR systems that are not inherently well-calibrated, calibration
involves training a model to transform raw outputs into interpretable
values on an appropriate scale. In weather forecasting, the calibration
model is trained on historical forecast data paired with observed
weather outcomes \citep{wilksExtendingLogisticRegression2009}. The model
learns parameters that transform raw predictions to better align with
real-world results. This enables the forecast to be interpreted in
probabilistic terms that can then be reliably used for decision making
- e.g.~``how likely is it to rain, and thus, should I carry an
umbrella?''

\subsection{Calibration in AV}\label{calibration-in-av}

As in weather forecasting, the outputs of AV systems are not always well-calibrated. An AV system may produce an output that resembles a Likelihood Ratio (LR), in
that it is comprised of similarity and typicality measures, but the
probabilities (\(P(E|H_p)\) and \(P(E|H_d)\)) may be inaccurate. 

This could reflect that even when an AV system is grounded in cognitive linguistic theory \citep{niniTheoryLinguisticIndividuality2023a}, the complexity and incomplete understanding
of individual language production constrain the AV model to simplify the underlying
generative mechanisms
\citep{argamonInterpretingBurrowssDelta2008, niniGrammarBehavioralBiometric2026}. If the assumptions of the model are not accurate, it is unlikely that the estimated probabilities will reflect the true probability of the observed evidence.

While uncalibrated LLRs may correctly order the strength of evidence, their
numerical magnitudes are not interpretable. This means that the values do
not reliably represent the likelihood of the evidence given each
hypothesis.

To transform the system's outputs into a calibrated LLR, a further process is therefore required. Most often, this is performed using logistic regression calibration \citetext{\citealp[
p.51]{ishiharaForensicAuthorshipClassification2011}; \citealp{morrisonTutorialLogisticregressionCalibration2013}}.

\subsection{Logistic Regression
Calibration}\label{logistic-regression-calibration}

\label{sec:logreg}

Logistic regression calibration was first established for speaker
recognition
\citep{gonzalez-rodriguezEmulatingDNARigorous2007, brummerLikelihoodratioCalibrationUsing2013, morrisonTutorialLogisticregressionCalibration2013}
and is now applied across a variety of forensic science disciplines
\citep{biosaEvaluationForensicData2020, macarullarodriguezImprovedLikelihoodRatios2024, ishiharaEstimatingStrengthAuthorship2022}.
In practice, one of the main challenges of logistic regression
calibration is acquiring and preparing the data needed to train a
logistic regression model.

\subsubsection{Data for Calibration}\label{data-for-calibration}

\label{sec:data}
The training data required for logistic regression calibration consists of outputs from the given forensic analysis system with accompanying ground-truth labels. In the context of AV, this requires a set of text pairs labelled as same-source (written by the same author) or different-source (written by different authors). It is necessary to have pairs of texts belonging to both source categories.

As language use varies across contexts \citep{biberRegisterPredictorLinguistic2012}, different conditions can lead to systematic variation in linguistic features, and consequently, AV scores. To ensure the logistic regression transformation generalises to the case data, the texts must therefore reflect the conditions of the case \citep{morrisonBiGaussianizedCalibrationLikelihood2024a}. However, the threshold of sufficient similarity to the case data is not clearly defined \citep{morrisonContextForensicCasework2021} and neither are the precise parameters that should be more closely aligned. The choices are therefore left to the expert, which, in turn, opens these choices to debate with opposing experts.

In practice, perfectly matched data can be difficult to obtain. For example, if the AV task concerns a malicious forensic text, like a threat, comparable texts may be rare or inaccessible \citep{niniRegisterVariationMalicious2017}. In such cases, controlling for additional variables, such as matching the dialect to the relevant population \citep{morrison2012},  may not be possible, whilst maintaining a sufficiently large dataset. 

The forensic linguist is therefore required to make decisions regarding how much and which data to use for calibration. To ensure that this subjective judgement is transparent and defensible \citep{morrisonContextForensicCasework2021}, these decisions must be validated through additional tests that verify the validity and reliability of the system \citep{morrison2012}. Validation tests add further complexity, as well as time and computational cost. 

Further time cost is incurred as the texts in the calibration dataset require preprocessing, whereby features that may introduce noise or bias into the analysis are removed. The specific preprocessing steps depend on the dataset and the AV system used and typically requires both automation and manual validation. As calibration data is often obtained from different sources to the case data, the steps required to achieve a standardised output may differ between the two.

Moreover, when pairing texts to create simulated authorship verification problems for the calibration dataset, experts must make a series of decisions. These decisions, while not arbitrary, may not be grounded in the existing literature due to limited research on the issue.

One example of this is the choice of anchoring method \citep{reindersSourceanchoredTraceanchoredGeneral2022}.  Anchoring pertains to the construction of the different-source comparisons in the calibration dataset. In AV, trace-anchoring is where the disputed text (\emph{Q}) is paired with texts from randomly selected alternative authors. In contrast, in source-anchoring, it is the known-author text (\emph{K}) that is paired with texts from randomly selected authors. Alternatively, the different-source problem set may be constructed only from authors that do not feature in the case data.

Importantly, each of these anchoring methods can result in systematically different LLR estimates \citep{heplerScorebasedLikelihoodRatios2012}. However, there is limited empirical validation regarding which anchoring strategy is optimal, particularly in the context of AV. Once again, the forensic linguistic expert must therefore validate such decisions for every case, through additional tests.

Once the pairs of texts are prepared, they are analysed using the selected AV model. For each pair of texts, the model may return a score that indicates both similarity and typicality but is not directly interpretable as a LLR. For the calibration dataset, each score is associated with a known ground-truth label (a probability of 1 if the
pair of text is same-source and 0 for different-source pairs).

Each step of this process must be explained and justified in the forensic report. This requires the forensic expert to demonstrate the validity of the procedure in a clear and accessible outline, that can be understood by juries or legal practitioners without specialist knowledge. Communicating all decisions clearly and transparently can be challenging, but is essential for maintaining the credibility of the analysis, and ensuring that the court can interpret the findings in an informed and appropriate way.

\subsubsection{Model Estimation}\label{model-estimation}

Once the calibration data is prepared it can be used to train a logistic regression model, which can subsequently be used to calculate a calibrated LLR. This section provides an overview of logistic regression model estimation. For a more in-depth tutorial see Morrison \citeyearpar{morrisonTutorialLogisticregressionCalibration2013}.

Logistic regression performs calculations in log-odds space, so the relationship between the uncalibrated score and the LLR is linear. The transformation is performed using Equation~\ref{eq-logreg}. where \(\alpha\) is the intercept and \(\beta\) is the slope.

\begin{equation}\phantomsection\label{eq-logreg}{
y = \alpha + \beta x}
\end{equation}

These parameters are estimated from the training dataset of same-source and different-source scores, and the resulting transformation can then be applied to the case data, where the source category is unknown. These scores are shifted by \(\alpha\) and scaled by \(\beta\) to obtain calibrated LLRs. 

The model is fitted using maximum likelihood estimation. To ensure that the calibration model outputs correspond to the LLR, equal class priors must be assumed.

Unlike generative models, which explicitly model the score distributions for same-source and different-source comparisons, logistic regression models the decision boundary between the two classes. This boundary follows a sigmoidal function, the position and steepness of which can be adjusted, but does not directly represent the underlying score distributions. Consequently, logistic regression calibration may be considered a relatively opaque approach.

\subsection{Calibration Evaluation}\label{calibration-evaluation}

To determine whether a logistic regression model has produced well-calibrated LLRs, it must undergo a validation test. The metric used to evaluate the validity of LLR systems is the cost of
log-likelihood ratio (\(C_{llr}\))
\citep{brummerApplicationindependentEvaluationSpeaker2006, morrisonMeasuringValidityReliability2011a}.
This metric is the established standard in the forensic analysis of
behavioural biometrics, with van Lierop et al. \citeyearpar[figure
1]{vanlieropOverviewLogLikelihood2024a} reporting its use in all
reviewed publications on stylometric analysis, and the majority of those
on speaker recognition.

To calculate the \(C_{llr}\), a validation dataset is required,
consisting of LRs for same- and different-source pairs of texts, with
ground-truth labels specifying whether each pair was written by the same
or different authors. Ideally, this validation dataset should resemble
the case conditions as closely as possible to ensure meaningful
performance evaluation. Consequently, the validation data is subject to
the same data-dependency limitations as the calibration data.

In the validation dataset, the ground-truth labels enable the
calculation of whether the AV model's prediction aligns with the true
observation. The \(C_{llr}\) is calculated using the following equation
\citep[eq. 3]{morrisonMeasuringValidityReliability2011a}:

\begin{equation}\phantomsection\label{eq-cllr}{
C_{llr} = \frac{1}{2}\left(\frac{1}{N_{s}} \sum_{i=1}^{N_{s}} \log_2 \Big(1 + \frac{1}{LR_{s_i}}\Big) + \frac{1}{N_{d}} \sum_{j=1}^{N_d} \log_2 \Big(1 + LR_{d_j}\Big)\right)
}\end{equation}

In this calculation, for same-author problems, a cost is incurred when
the LR (\(LR_s\)) fails to strongly support the prosecution hypothesis
(\(1 + 1/LR_s\)). This cost decreases as evidence in favour of the
correct hypothesis increases. Conversely, the term \((1 + LR_d)\)
penalises LRs that incorrectly favour the prosecution hypothesis. The
logarithmic transformation converts LRs into interpretable, additive
quantities. Averaging over both trial types (\(1/N_s\) and \(1/N_d\))
and weighting them equally yields a single scalar measure that jointly
reflects discrimination and calibration quality.

The lower the \(C_{llr}\), the better the performance of the AV
system. However, whilst it is established that a \(C_{llr}\) of 1
indicates a system is uninformative, it is not well-established what a
good \(C_{llr}\) is \citep{vanlieropOverviewLogLikelihood2024a}.

\section{Authorship Verification}\label{authorship-verification-with-lambdag}

\label{sec:av}

\subsection{Properties of Authorship Verification Methods}\label{av-cat}

AV methods can be categorised into three distinct groups: unary, binary intrinsic and binary extrinsic methods \citep{halvaniAssessingApplicabilityAuthorship2019}. The categories are defined by the data required and how this data is utilised to establish a decision criteria for AV.  

A unary AV method \cite[e.g.,][]{ noeckerjrDistractorlessAuthorshipVerification2012, halvaniAuthorshipVerificationAbsence2018} only requires the case data in order to establish its decision criteria. The case data refers to the questioned text (\(Q\)) and sample text(s) known to be written by the candidate author (\(K\)). 

As a result of this characteristic data requirement, it is an intrinsic property of unary AV methods that only one class is considered. As such, a LLR cannot be obtained from a unary AV method, as, by definition (Equation~\ref{eq-LLR}), the probability of the evidence if the candidate author did not write the questioned text, must also be considered.   

Comparatively, a binary method utilises data that is external to the case data, allowing for LLR estimation. Binary intrinsic methods \cite[e.g.,][]{ halvaniUsefulnessCompressionModels2017, boenninghoffSimilarityLearningAuthorship2019, huertas-tatoUnderstandingWritingStyle2024} use training data to learn how to distinguish between same-author and different-author classes. Comparatively, binary extrinsic methods \cite[e.g.,][]{koppelDeterminingIfTwo2014a, pothaImprovedImpostorsMethod2017a, niniGrammarBehavioralBiometric2026} compare the language in external reference documents \(\mathbb{R}\), to the language used in \(Q\). This provides a point of comparison for the similarities found between \(Q\) and \(K\).

Table \ref{tbl-comp} demonstrates that across these three categories, the existing methods share a particular property: that meaningful likelihood ratios cannot be calculated without a separate calibration model, if at all. The present research proposes two novel approaches that do not share this property, creating a new class of authorship verification method. 

\begingroup\fontsize{8}{10}\selectfont

\begin{longtable}[t]{>{\raggedright\arraybackslash}p{1.75cm}>{\centering\arraybackslash}p{1.5cm}>{\centering\arraybackslash}p{1.75cm}>{\centering\arraybackslash}p{1.5cm}>{\centering\arraybackslash}p{1.75cm}>{\centering\arraybackslash}p{1.75cm}}

\caption{\label{tbl-comp}Comparative overview of 
AV categories,
highlighting data requirements}

\tabularnewline

\toprule
Method & Calibration Data Required & Reference Corpus Required \\
\midrule
Proposed Methods & No & Yes\\
Unary & N/A & No \\
Binary Intrinsic & Yes & No \\
Binary Extrinsic & Yes & Yes \\
\bottomrule

\end{longtable}

\endgroup{}

The proposed methods constitute an adjustment of an existing binary extrinsic method. By eliminating the need for a separate calibration model, this, in turn, reduces both
the data requirement and the complexity of the method.

\subsection{LambdaG}\label{lg}

The binary-extrinsic method employed in this paper is LambdaG \citep{niniGrammarBehavioralBiometric2026}, which relies on n-gram language modelling. This method has a unique property that establishes it as a suitable focus for this study. That is, it produces scores intrinsically formulated as LLRs, albeit uncalibrated prior to the application of logistic regression calibration. Therefore, whilst LambdaG currently requires post-hoc calibration, its formulation may allow for alternative, simpler transformations to achieve a calibrated LLR output, which would not be possible with any other existing AV method.

To apply the method, first the input data (\(Q\), \(K\), and the set of reference texts $\mathbb{R}$) must be \textit{content masked}. This is a computational procedure which accounts for the influence of topic on authorship analysis by obscuring words that carry lexical meaning, as opposed to serving a grammatical function. \citet{niniGrammarBehavioralBiometric2026} employ the \textit{POS-Noise} method of content masking \citep{halvaniPOSNoiseEffectiveCountermeasure2021a}, which substitutes tokens that contain topical information with their part-of-speech. For example, the sentence \emph{the cow jumped over the moon} becomes \emph{the N V over the N}. The tokens deemed topic-agnostic are identified using a pre-defined list \footnote{The list can be found here: https://bit.ly/ARES-2021 \citep[p.3]{halvaniPOSNoiseEffectiveCountermeasure2021a}}.

The method then consists in calculating $\lambda_G$, the ratio of the the likelihood of each token in context given a \textit{grammar model} of the author of $K$, $G_K$, and the likelihood of each token in context given a set of grammar models from the reference population, $G_R = \{G_1, G_2, ... G_r\}$.

To estimate these grammar models, each of the content masked texts across the questioned, candidate and reference data are tokenised into sentences. The set of sentences in the content masked \(K\) (\(\mathbb{S}_K\)) are then used to train an n-gram grammar model (\(G_K\)).

An n-gram language model estimates the conditional probability of a word by applying the Markov assumption of order \citep[p.40]{jurafskySpeechLanguageProcessing2026}. In the context of text generation, a first-order Markov assumption states that the probability of the token at position \(j\) in a sentence depends only on the token itself. Higher order Markov assumptions, e.g.~\(N = 10\), allow more context into the n-gram language model. The probability of a word sequence of length $n$ (\(t_1 ... t_n\)) using a 10-gram language model would be calculated using Equation~\ref{eq-ngram}. For the first nine tokens, beginning of sentence tokens (\([BOS]\)) are prepended.

\begin{equation}
\label{eq-ngram}
P(t_1 ... t_n) \approx \prod_{j=1}^{n} P(t_j \mid t_{j-9}, t_{j-8}, \dots, t_{j-1})
\end{equation}

N-gram languages models estimate probabilities from corpus counts using
maximum likelihood estimation. For LambdaG, these probabilities are then adjusted using Kneser-Ney smoothing \citep{kneserImprovedBackingoffMgram1995, chenEmpiricalStudySmoothing1999}. Kneser-Ney smoothing redistributes probability mass by backing off to lower-order distributions when higher-order n-grams are sparse or unseen. The probabilities of tokens are weighted according to the number
of distinct contexts in which they appear.

To train the reference grammar models, sentences are randomly sampled from \(\mathbb{R}\) and concatenated. This produces a set of sentences written by multiple different authors. The number of sentences sampled is equal to the number of sentences in \(K\). On iteration \(i\) this set of sentences is used to train \(G_i\). This process of sampling and training is repeated \(r\) number of times. 

For each token (\(t_j\)), in each sentence, the \(\lambda_G\) is calculated by computing the probability that \(t_j\) is the next token, conditioned on all preceding tokens in that sentence (\(t_{<j}\)) as Equation~\ref{eq-lambdagdef}. 

\begin{equation}
\label{eq-lambdagdef}
\lambda_G(t_j|t_{<j}) = \frac{1}{r} \sum_{i=1}^{r} \log \frac{P(t_j|t_{<j} ; G_k)}{P(t_j|t_{<j} ; G_i)}
\end{equation}

The probability of \(t\) given \(G_K\) constitutes a similarity measure between \(Q\) and the texts known to be written by the candidate author (\(K\)). The mean probability of \(t\) given the reference grammar models is a typicality measure capturing distributional norms across language users. By calculating the ratio of these probabilities to produce a \(\lambda_G\) value, that value is inherently formulated as a LLR. 

The final score for the questioned text is the sum of the $\lambda_G$ scores for each token in the $Q$ text (Equation~\ref{eq-lambdagfinal}).

\begin{equation}
\label{eq-lambdagfinal}
\lambda_G(Q) = \sum{\lambda_G(t_j|t_{<j})}
\end{equation}

The full algorithm is provided in \ref{ap-a}.

To illustrate this process, consider a content masked \(Q\) that contains the sentence \emph{The N V over the N}. The \(\lambda_G\) value of the token \emph{over} is the logarithm of the probability that \emph{over} is the next token in the sequence \emph{The N V} given \(G_K\), divided by the mean probability that \emph{over} is the next token in the sequence according to the \(r\) grammars in \(G_{R}\).  The process is repeated for every token in every sentence in \(Q\). The \(\lambda_G\) for each token is added to produce an final overall \(\lambda_G\) for the text.

Beyond its LLR formulation and state-of-the-art performance, LambdaG also presents a number of other advantages. These include that it is designed to exhibit minimal sensitivity to topic variation, performs consistently across registers, including very short texts, and is grounded in cognitive linguistic theories.

The following section will outline that despite \(\lambda_G\) being formulated as a LLR, it does not reliably reflect the true strength of the linguistic evidence. It is therefore not a meaningful LLR, it is an uncalibrated score. This may indicate that, despite being grounded in valid linguistic assumptions, LambdaG does not model language productivity with complete accuracy. As a result, like other AV methods, LambdaG requires further calibration to obtain a calibrated LLR (\(\Lambda_G\)). \citet{niniGrammarBehavioralBiometric2026} achieve this using logistic regression calibration.

\section{Score Inflation and
Corrections}\label{score-inflation-and-corrections}

\label {sec:propm}

We propose that, in the case of LambdaG, calibration is necessary because, as LambdaG sums independent token-level contributions, it adopts the naïve Bayes assumption of conditional independence between long-distance input features. This simplifying assumption introduces a systematic weakness affecting all naïve Bayes algorithms, whereby the reliability of the resulting probability may be compromised.

We hypothesise that the issue is the inherent repetition in natural language production. As token-level contributions are aggregated, repeated evidence compounds. However, it is plausible to assume that subsequent repetitions of the same pattern contribute less evidential value than initial occurrences. This may lead to the strength of authorship evidence being overstated in longer or highly repetitive texts, with repeated stylistic features contributing disproportionately to the overall score. The following section outlines the supporting evidence that grounds this hypothesis. 

\subsection{Distribution of Raw
Scores}\label{distribution-of-raw-scores}

\begin{figure}

\centering{

\includegraphics[width=1\linewidth,height=\textheight,keepaspectratio]{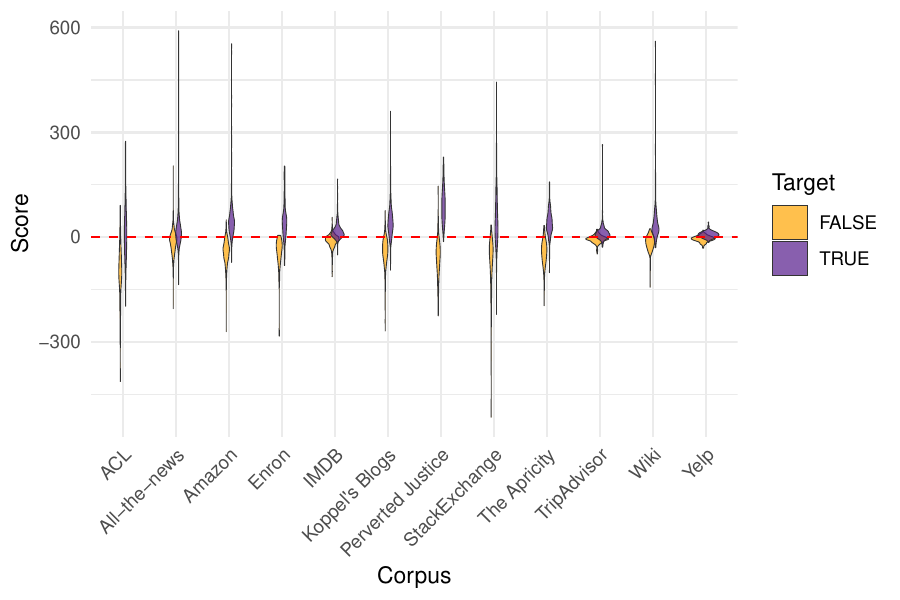}

}

\caption{\label{fig-dist}Distribution of raw LambdaG scores for each
corpus adopted from Nini et al.~(2026), shown as violin plots. Colours
indicate target label.}

\end{figure}

Examining the \(\lambda_G\) distributions for same-source and different-source problems demonstrates substantial overestimation of
evidential strength. Figure~\ref{fig-dist} presents these distributions
across the corpora used to evaluate LambdaG in
\citet{niniGrammarBehavioralBiometric2026}. \(\lambda_G\) values greater
than 600 were omitted from Figure~\ref{fig-dist} for visualisation
purposes. All excluded values, the most extreme of which was 3,321.84,
correspond to the \textit{All-the-news} corpus. These values were retained
for all later analysis.

A LLR of 4, for example, indicates that the observed
evidence is 10,000 times more likely under the prosecution hypothesis
than under the defence hypothesis. However, even with the most extreme
values omitted, the remaining \(\lambda_G\)
values in Figure~\ref{fig-dist} still span a notably large range.
Moreover, aside from the remaining extreme values in the tails of the
distribution, many values still exceed four, with numerous values
between 50 and 100.  \citet[p.15]{niniGrammarBehavioralBiometric2026} demonstrate that these overinflated raw scores are misleading, with \(\lambda_G\) yielding a \(C_{llr}\) of greater than one across each corpus. 

Figure~\ref{fig-dist} also illustrates that, across most corpora, the distributions of \(\lambda_G\) for same-source texts and different-source texts are approximately sign-aligned. True problems tend to obtain a positive \(\lambda_G\) value and false problems a negative \(\lambda_G\). Given that \(\lambda_G\) does not need to be shifted, and is already constructed as a LLR, the primary role of calibration may therefore be to scale the scores to mitigate overstated evidence. Logistic regression calibration may therefore constitute an excessively complex solution for what might be solved by a simpler transformation, such as normalisation.

\subsection{The Square Root
Correction}\label{the-square-root-correction}

The first innovative approach, the Square Root Correction (Equation
\ref{eq-sqr}), scales \(\lambda_G\) using a simple metric based on the number of tokens in the questioned document, \(N(Q)\). The longer a text is the more likely it is to contain repetitions \citep{herdanTypetokenMathematics1960, heapsInformationRetrievalComputational1978a}. Therefore, as text length increases, additional tokens are increasingly likely to present evidence that has already been seen. 
Subsequent repetitions of the same pattern are unlikely to contribute fully independent evidential weight, as once a stylistic feature has been observed, its recurrence becomes increasingly expected
\citep{coulthardAuthorIdentificationIdiolect2004, niniTheoryLinguisticIndividuality2023a}. As a result, with increasing tokens, the average contribution of evidential strength for each token is likely to diminish. 

\begin{equation}\phantomsection\label{eq-sqr}{ \Lambda_G = \frac{\lambda_G}{\sqrt{N(Q)}} }\end{equation}

The selection of a square-root transformation is consistent with
statistical principles, where variance often grows approximately
proportionally with the aggregation of weakly-dependent input features,
causing dispersion to increase approximately proportionally to
\(\sqrt{N(Q)}\). Similar forms of square-root scaling are also used in
other domains, such as the scaling factor applied in attention
mechanisms within machine learning \citep{vaswaniAttentionAllYou2017a}.

Normalising solely by text length would implicitly assume that every token contributes equal evidential value. In contrast, transforming a score by the square root of text length, captures the diminishing contributions from additional tokens. The Square Root Correction can therefore moderate the inflated predictions of evidential strength for longer texts, without excessively penalising them.

\subsection{Adversarial Manipulation}\label{adversarial-manipulation}

While the Square Root Correction mitigates score inflation, it does not
directly address the underlying independence assumption that may cause
such inflation. To more directly account for repeated tokens providing
increasingly redundant stylistic evidence, normalisation must
incorporate not only text length but also a measure of repetition, or
information redundancy.

One scenario where scaling by length is likely not sufficient to address
the effect of repeated tokens is adversarial manipulation of AV systems,
for instance, when impersonation is attempted. If an adversary is aware of a specific linguistic preference of the individual they are impersonating, then they would likely repeat this known feature several times.

One example of this is the Starbuck case, where Jamie Starbuck attempted to impersonate his wife, Debbie \citep{grantStarbuckCase2022}. Debbie used a high-frequency of semi-colons, and having noticed this, when impersonating Debbie, Jamie employed a high-frequency of semi-colons comparative to his own typical usage. The feature was repeated to such an extent that the frequency of semi-colons in the impersonated texts was higher than had been observed in texts known to be written by Debbie \citep{ForensicAuthorshipAnalysis}. 

With the current LambdaG system, such impersonation may have the desired effect,
because each occurrence would independently increase the score. The
model could not differentiate this artificial inflation from natural
language. As a result, the repeated use of a deliberately imitated
stylistic feature may artificially dominate the overall \(\lambda_G\) of
the text. This means the AV system is vulnerable to manipulation,
reducing robustness in forensic settings.

\subsection{The Hapax Correction}\label{the-hapax-correction}

The second proposed correction, the Hapax Correction (Equation
\ref{eq-hc}), aims to mitigate inflated evidential strength in instances
of limited stylistic diversity, relative to the length of the text. To
capture the degree of repetition and redundancy in a text, the
correction incorporates the number of \textit{hapax legomena}, \(V_1(Q)\) - the number of
distinct types that only appear once in a text, in
this case, the questioned document.

\begin{equation}\phantomsection\label{eq-hc}{ 
\Lambda_G = \lambda_G \times \frac{V_1(Q)}{N(Q)}
}\end{equation}

For this correction, hapax legomena are counted in the content-masked
text rather than on the original text, as the masked representation
constitutes the actual input to the LambdaG model. The aim of the
normalisation process is to counteract the accumulation of scores
arising from repeated encounters with the same tokens, including the
parts-of-speech introduced through POS-Noise. Using masked token counts
ensures that normalisation operates over the same set of tokens that
contributes to score accumulation, thereby maintaining consistency
between the source of bias and the corrective mechanisms.

As raw counts of hapax legomena are not comparable for texts of
different lengths, the Hapax Correction calculates the ratio of hapax
legomena to the total token count of the questioned document (\(N\)).
This provides a more reliable measure of linguistic productivity and
lexical diversity. As this value is always between 0 and 1, multiplying
\(\lambda_G\) by this ratio proportionally scales texts containing a
high degree of repetition. For texts with relatively few hapax legomena,
and therefore a greater proportion of repeated tokens, the score is
reduced more than for a text with a wide range of stylistic evidence.
This aims to reduce inflated evidential strength resulting from
redundant or highly correlated linguistic patterns.

\section{Methodology}\label{methodology}

To address the question of whether \(\Lambda_G\) can be derived from
\(\lambda_G\) without training a calibration model, we evaluated the
Hapax Correction and Square Root Correction against logistic regression
calibration across fifteen corpora. For a subset of these corpora, the
corrections were tested across a range of text length configurations,
from 100-9,500 tokens for the Q and K texts. The data preparation and
analysis were executed using R. In particular, the analysis relied on
the \textit{idiolect} package \citep{niniIdiolectPackageForensic}.

\subsection{Data}\label{data}

\subsubsection{Overview}\label{overview}

The data used in this study comprises fifteen different corpora,
alongside corresponding problem sets. Twelve of the corpora were adopted
from \citet{niniGrammarBehavioralBiometric2026}. These consisted of
academic papers (\textit{ACL}), newspaper articles (\textit{All-the-news}),
blog posts (\textit{Koppel's Blogs}), four corpora of reviews
(\textit{Amazon}, \textit{IMDB}, \textit{TripAdvisor} and \textit{Yelp}),
Wikipedia talk pages (\textit{Wiki}), two forum corpora (\textit{The
Apricity}) and (\textit{StackExchange}) , emails (\textit{Enron}) and chat
logs (\textit{Perverted Justice}).

The component corpora represent a range of scenarios encountered in
real-word AV cases. Some of the complications captured by the dataset
include very short texts (\textit{Perverted Justice} and \textit{Yelp}),
texts of varying lengths (\textit{Stack Exchange}), very formal texts
(\textit{ACL}), texts with a high frequency of non-standard lexical items
(\textit{Perverted Justice} and \textit{All-the-news}), texts written by
the same author several years apart (\textit{Yelp} and \textit{ACL}), and
texts written by the same author on different topics and different
authors on the same topic (\textit{Stack Exchange}). Further information
on each of these corpora, their preparation, and challenges is detailed
by \citet[Section 6.1]{niniGrammarBehavioralBiometric2026}.

Three further corpora were compiled to complement the existing datasets,
introducing higher volumes of data per author and previously
unrepresented registers. The two registers on which LambdaG has not
previously undergone formal evaluation are tweets (\textit{Twitter}) and
text messages (\textit{Bolt}). The register of the final corpus employed
in this paper comprises emails (\textit{Avocado}). This register has been
examined before (the \textit{Enron} corpus), but only with substantially
smaller amounts of data for each author (averaging 876 tokens of
disputed data). This analysis of \textit{Avocado} and \textit{Twitter}
implements approximately 9,500 tokens of disputed data per author,
enabling the assessment of the proposed corrections, and logistic
regression calibration, across a range of text-length configurations.

Authors in each corpus were randomly assigned to either the case data or
the calibration dataset. The case data represents the target AV problems
and is the data on which predictions are made, whereas the calibration
data is used solely for training the logistic regression model
implemented in the baseline approach. For the corpora adopted from
\citet{niniGrammarBehavioralBiometric2026}, 40\% of each corpus was used
for calibration data, and the remaining 60\% for case data.
Comparatively, for the three new corpora a 50/50 split was implemented.
Each author features in two problems, one same-source pairing and one
different-source pairing. In total, there are 13,146 problems. 4,766 of
the problems are calibration data and 8,380 are case data, both are
evenly split between same-source and different-source pairs.

\subsubsection{Preparation}\label{preparation}

The data preparation processes described below were designed to mirror the standard that would be applied to real-world case data. However, no standardised framework exists, as preprocessing requirements differ depending on each register and corpus. 

Two of the corpora adopted from \citet{niniGrammarBehavioralBiometric2026} underwent additional
preprocessing. In \textit{All-the-news}, duplicate sentences
within the same text were removed, as some texts contained repeated
sections. In \textit{Enron}, emails were removed if they
contained a sequence of 10 or more tokens that also featured in another
text, to reduce duplication. To address outliers in token distributions,
an inter-quartile range-based filter \citep{tukeyExploratoryDataAnalysis1977} with a multiplier of 0.8 was applied. This multiplier was selected on the basis of manual validation of the output. Authors below
the lower threshold were removed, while authors above the upper
threshold were trimmed to within the accepted range.

The first of the new corpora, \textit{Avocado}, consists of 64,858 emails
from \(40\) different accounts. The data originates from the Avocado
Research Email Collection \citep{oarddouglasAvocadoResearchEmail2015}, a
corpus of emails from a defunct informational technology company. The objective of preprocessing \textit{Avocado} was to emulate the initial preprocessing of \textit{Enron} \citep[p.22]{niniGrammarBehavioralBiometric2026}. However, as \textit{Enron} is much smaller, some of this was achieved manually. For \textit{Avocado} our aim was to automate as much of the process as possible. 

 That being said, a small number of accounts identified as non-human were manually excluded.
As part of automated preprocessing, salutations and signatures were removed. Duplicate and near-duplicate texts were identified using MinHash LSH
\citep{mullenMinhashLocalitysensitiveHashing2020} (\(240\) minhashes, \(80\) bands, \(10\)-token n-grams); based on manual inspection, any
pair with non-zero Jaccard similarity was removed. Automated preprocessing also
included lowercasing, standardising elongated words, replacing email
addresses with a placeholder, removing URLs and non-ASCII characters,
trimming whitespace, and removing texts shorter than three tokens or
longer than 1,000 tokens. Each author in \textit{Avocado} has a minimum
of 100,000 tokens, of which approximately 35,000 texts are labelled as
\emph{unknown} (Q data) This was achieved by randomly assigning texts
until the token threshold was exceeded.

The second new corpus, \textit{Twitter}, consists of 576,602 tweets by
\(60\) authors. From the original dataset
\citep{grieveFrontiersMappingLexical2019}, a sample of authors with more
than 135,000 tokens was taken. Authors were manually reviewed to exclude
automated accounts. Duplicates were removed, as were tweets beginning
\emph{rt :}, as it indicated a retweet. Within texts, emojis, leading
punctuation, non-printing Unicode characters and redundant white space
were also removed. URLs were replaced with \emph{<url>}, and tags were replaced with
\emph{userid}. Elongated words were normalised, and consecutive
sequences of more than three identical punctuation marks were reduced to
three. Any texts that now had less than three tokens were then removed.
The total word count for each author was then calculated, and this
preprocessing was repeated until a set of \(60\) authors with greater
than 135,000 tokens was obtained. Approximately 35,000 tokens were
allocated to the \emph{unknown} data for each author, and 100,000 tokens
to the known (\emph{K}) data.

Lastly, the \textit{Bolt} Corpus comprises 3716 messages authored across
\(46\) individuals. The data consists of natural text (SMS) and chat
messages sourced from the BOLT (Broad Operational Language Translation)
English SMS/Chat corpus \citep{chensongBOLTEnglishSMS2018}. The
preprocessing of this dataset involved removing URLs, non-printing
Unicode characters, redundant internal and trailing whitespace and
non-ASCII characters, as well as normalising word elongations. Texts
with fewer than three tokens, and texts from authors with fewer than
\(1,000\) aggregated tokens were removed from the corpus. For each
author, texts were randomly sampled and assigned as \emph{unknown} until
the \(500\) token threshold was exceeded. The same threshold and process
was applied for known texts. Remaining texts were also removed.

\subsection{Protocol}\label{protocol}

LambdaG was applied to each corpus following the workflow outlined by
\citet{niniGrammarBehavioralBiometric2026}. To prepare the texts for
input, content masking was applied using the POSNoise method with the
spaCyR part-of-speech tagger (model: \texttt{en\_core\_web\_sm})
\citep{halvaniPOSNoiseEffectiveCountermeasure2021a, benoitSpacyrWrapperSpaCy2023a, niniIdiolectPackageForensic2024a}.
The texts were also tokenised into sentences using spaCyR (model:
\texttt{en\_core\_web\_sm})
\citep{benoitSpacyrWrapperSpaCy2023a, niniIdiolectPackageForensic2024a}.

Two hyperparameters are defined within LambdaG: \(r\) (the number of
iterations, i.e.~the number of reference grammar models) and \(N\) (the
order of the model) \citep{niniGrammarBehavioralBiometric2026} . Both
hyperparameters were set to the default values provided in the Idiolect
package: \(r = 30\) and \(N = 10\)
\citep{niniIdiolectPackageForensic2024a}. Nini et al. \citeyearpar[Figure
3]{niniGrammarBehavioralBiometric2026} found LambdaG is robust to changes in these hyperparameters.

On the twelve corpora adopted from
\citet{niniGrammarBehavioralBiometric2026}, LambdaG was applied to the
full texts. On \textit{Avocado}, \textit{Twitter} and \textit{Bolt} the
method was evaluated across a range of text length conditions. A
condition refers to the combination of \emph{Q} length and \emph{K}
length. For each of the three corpora, all texts were content masked and
tokenised into sentences, before they were concatenated into \emph{Q}
and \emph{K} data respectively. This ensured that text boundaries were
maintained in sentence tokenization.

From the \emph{Q} data (the concatenated list of tokenised sentences
labelled \emph{unknown}), sentences were randomly sampled until the
cumulative token count exceeded the specified \emph{Q} length. The same
procedure was applied to the \emph{K} data, using the corresponding
\emph{K} length as the threshold. LambdaG was then run on each problem
in turn. This process was applied to both the case and the calibration
data.

As a baseline, raw \(\lambda_G\) scores were calibrated using logistic
regression calibration, following the approach by
\citet{niniGrammarBehavioralBiometric2026}. The \(\lambda_G\) values
calculated for each same-source and different-source problem in the
calibration dataset were used to train the logistic regression model.
The model estimates the optimal parameters to transform \(\lambda_G\)
values into \(\Lambda_G\) by aligning predicted probabilities with
observed outcomes.

The proposed corrections (the Square Root Correction (Equation
\ref{eq-sqr}) and the Hapax Correction (Equation \ref{eq-hc})) provide an
alternative approach to approximate these optimal affine
transformations. These are also applied to \(\lambda_G\), enabling the
production of \(\Lambda_G\) without training a logistic regression
model.

Performance was evaluated using \(C_{llr}\). As LambdaG is a stochastic
method, each analysis was repeated five times, and the mean \(C_{llr}\)
was reported.

\section{Results}\label{results}
\label{sec:res}
\subsection{Performance Overview}\label{performance-overview}

\label{sec:po}

Table \ref{tbl-overview} provides an overview of the results across all
fifteen corpora, by outlining the win rate and close loss rate for each
pairwise comparison among the three methods. Each corpus was weighted
equally, thus where multiple tests were conducted on a corpus, the win
and close loss percentages were first calculated at the corpus level and
then incorporated into an overall average across all corpora. 

The threshold for close loss rate was defined as within 5\% of the comparator. A relative threshold was chosen to provide a
consistent and interpretable definition of comparable performance across performance results of varying magnitudes. Support for the threshold of 5\% will be outlined in Table \ref{tbl-development}, which indicates performance differences within this parameter may be attributable to noise. 

\begingroup\fontsize{8}{10}\selectfont

\begin{longtable}[t]{>{\raggedright\arraybackslash}p{2.2cm}>{\centering\arraybackslash}p{0.8cm}>{\centering\arraybackslash}p{0.8cm}>{\centering\arraybackslash}p{0.8cm}>{\centering\arraybackslash}p{0.8cm}>{\centering\arraybackslash}p{0.8cm}c}

\caption{\label{tbl-overview}Pairwise performance comparison of methods
across corpora. Columns show each methods performance against the
comparator, with each corpus weighted equally. Metrics include the
average percentage of tests where the method performed better or equal
(lower values indicate better performance) and the proportion of losses
that were within 5\% of the comparator}

\tabularnewline

\toprule
\multicolumn{1}{c}{ } & \multicolumn{2}{c}{Log Reg vs Square Root} & \multicolumn{2}{c}{Log Reg vs Hapax} & \multicolumn{2}{c}{Square Root vs Hapax} \\
\cmidrule(l{3pt}r{3pt}){2-3} \cmidrule(l{3pt}r{3pt}){4-5} \cmidrule(l{3pt}r{3pt}){6-7}
Metric & Log Reg & Square Root & Log Reg & Hapax & Square Root & Hapax\\
\midrule
Cases outperforms comparator (\%) & 74.80 & 25.20 & 54.60 & 45.40 & 21.80 & 78.20\\
Losses within 5\% of comparator (\%) & 40.18 & 21.36 & 49.13 & 56.91 & 23.89 & 45.29\\
\bottomrule

\end{longtable}

\endgroup{}

On average across the corpora, the Hapax Correction outperformed the
performance of logistic regression calibration in 45.4\% of tests. Where
the Hapax Correction was outperformed by logistic regression
calibration, its performance remained within 5\% of logistic regression
calibration in a higher proportion of tests on average (56.91\%) than
the reverse comparison (49.13\%). The Square Root Correction was
outperformed by each of the other approaches in over 70\% of tests (on
average). The close loss rate was 21.36\% when compared to logistic
regression, and 23.89\% when compared to the Hapax Correction.

The following sections will provide a more fine-grained view of this analysis. We will detail the comparative performance of the three approaches by corpus and across text length conditions. 

\subsection{Performance by Corpus}\label{sec:by-corpus}

Table~\ref{tbl-development} shows the performance of LambdaG on the
twelve corpora adopted from \citet{niniGrammarBehavioralBiometric2026},
when transformed using logistic regression calibration, the Square Root Correction and the Hapax Correction. Performance is measured using \(C_{llr}\). Each analysis was run five times and the mean and standard deviation was calculated. The results are separated by corpus, for each approach.

\begingroup\fontsize{10}{12}\selectfont

\begin{longtable}[t]{lccc}

\caption{\label{tbl-development}Performance evaluation of logistic
regression calibration, the Square Root Correction and the Hapax
Correction across twelve corpora, measured using Cllr. Results are
reported as mean Cllr with standard deviation across five random seeds.}

\tabularnewline

\toprule
Corpus & LogReg & SquareRoot & Hapax\\
\midrule
ACL & 0.888 ± 0.003 & 0.756 ± 0.008 & 0.777 ± 0.010\\
All-the-news & 0.808 ± 0.003 & 0.804 ± 0.003 & 0.856 ± 0.006\\
Amazon & 0.302 ± 0.001 & 0.437 ± 0.001 & 0.313 ± 0.001\\
Enron & 0.518 ± 0.026 & 0.505 ± 0.007 & 0.479 ± 0.014\\
IMDB & 0.658 ± 0.002 & 0.706 ± 0.002 & 0.657 ± 0.006\\
\addlinespace
Koppel's Blogs & 0.412 ± 0.001 & 0.486 ± 0.001 & 0.409 ± 0.001\\
Perverted Justice & 0.219 ± 0.002 & 0.253 ± 0.002 & 0.226 ± 0.003\\
StackExchange & 0.509 ± 0.011 & 0.533 ± 0.008 & 0.508 ± 0.012\\
The Apricity & 0.369 ± 0.009 & 0.451 ± 0.002 & 0.354 ± 0.007\\
TripAdvisor & 0.761 ± 0.004 & 0.783 ± 0.003 & 0.780 ± 0.008\\
\addlinespace
Wiki & 0.459 ± 0.013 & 0.579 ± 0.004 & 0.474 ± 0.011\\
Yelp & 0.713 ± 0.004 & 0.783 ± 0.001 & 0.739 ± 0.004\\
\bottomrule

\end{longtable}

\endgroup{}

The logistic regression calibration results reproduce the results of \citet[p.15]{niniGrammarBehavioralBiometric2026} Discounting the two datasets for which additional preprocessing was applied (\textit{Enron} and \textit{All-the-news}), the mean absolute difference across all datasets was 0.012 (rounded to three decimal places; all comparisons also used \(C_{llr}\) values rounded to three decimal places). This demonstrates high consistency between the present study and \citet{niniGrammarBehavioralBiometric2026}, and highlights the reproducibility of LambdaG.  

No single approach has superior performance across all corpora, and the
three approaches demonstrate similar patterns of performance across each
register, for example, all three achieve their lowest \(C_{llr}\) values
on the \textit{Perverted Justice} corpus (\(0.219\), \(0.253\) and
\(0.226\)).

Across all three approaches, the low standard deviation values indicate
that performance is stable across runs. This suggests low sensitivity to
random seed initialisation. Of the approaches, logistic regression
calibration shows slightly higher variability than the proposed
corrections.

The corpus with the greatest standard deviation was \textit{Enron}, with a standard deviation of 0.026 under the logistic regression calibration condition. This constitutes approximately 5\% of the mean \(C_{llr}\) obtained for that condition. This justified the 5\% threshold used in Table \ref{tbl-overview}, as differences within 5\% of the mean \(C_{llr}\) across conditions may be comparable to intrinsic variability observed within conditions.

\subsection{Text Length Analysis}\label{text-length-analysis}

\subsubsection{\textit{Avocado}}\label{c_avocado}

\begin{landscape}
\begin{figure}

\centering{

\includegraphics[width=1\linewidth,height=\textheight,keepaspectratio, trim = 0 3cm 0 3cm, clip]{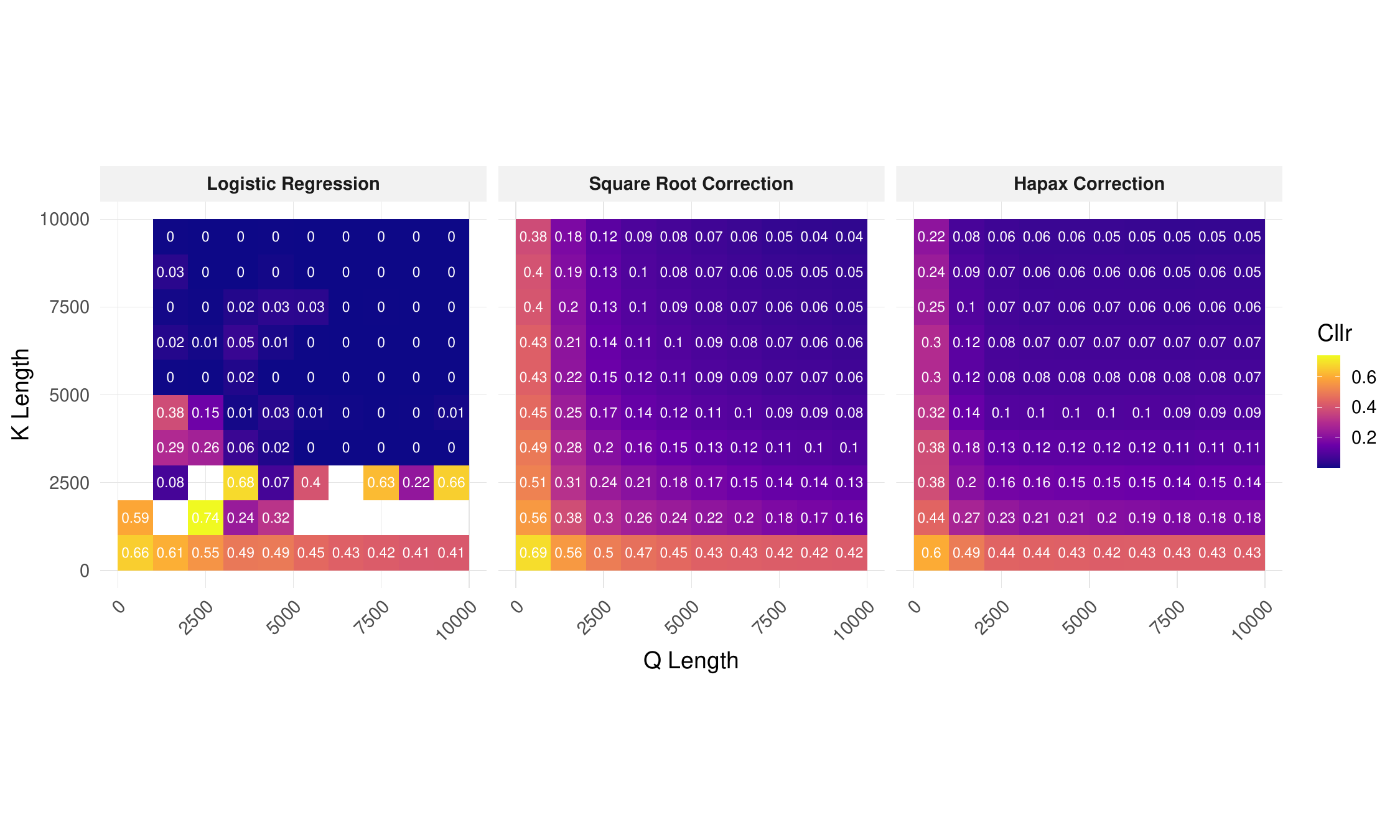}

}

\caption{\label{fig-avo}Comparison of three approaches on the Avocado
dataset showing the relationship between Q Length, K Length and mean
Cllr. Each panel corresponds to a post-processing method applied to
LambdaG scores: logistic regression calibration (left), the Square Root Correction (middle) and the Hapax
Correction (right).}.

\end{figure}
\end{landscape}

Figure \ref{fig-avo} presents the results obtained from the analysis of
\textit{Avocado}. It shows how the length of the disputed document
(\emph{Q}) and the known-author data (\emph{K}), impacts \(C_{llr}\)
across the three different approaches. The \emph{Q} and \emph{K} lengths
ranged from approximately 500 to 9,500 tokens, sampled at 1,000 token
intervals. Every possible pairwise combination of \emph{Q} and \emph{K}
length was evaluated. The mean \(C_{llr}\) is represented by the colour
gradient, with empty cells representing a \(C_{llr}\) \textgreater{} 1,
i.e.~misleading output \citep{vanlieropOverviewLogLikelihood2024a}. For
the retained datapoints the mean \(C_{llr}\) values are provided within
the heatmap and rounded to two decimal places; therefore, where a mean
\(C_{llr}\) is shown as \(0\), this indicates that the true value was
less than 0.005.

Figure~\ref{fig-avo} demonstrates that both proposed corrections yield a
positive performance, with the lowest observed \(C_{llr}\) of 0.05 for
the Hapax Correction and as low as 0.04 for the Square Root Correction -
both achieved when \emph{Q} and \emph{K} were at their longest tested
length (9,500). However, logistic regression calibration achieved the
best peak performance, with values \(<\) 0.005 in certain instances
where \emph{K} length \(\ge\) 3,500.

For conditions where the logistic regression calibration obtained a
\(C_{llr}\) \textgreater 0.1, the Hapax Correction performed better or
equal in 92.3\% of cases, and the Square Root Correction in 84.6\% of
instances; in a further 7.69\% and 10.3\% of cases respectively, the
proposed methods were within 5\% of the baseline \(C_{llr}\).

When logistic regression calibration achieved a \(C_{llr}\) \(\le\) 0.1,
percentage difference is less informative due to small values. In these
cases, the maximum observed difference was 0.13 for the Hapax Correction
and 0.23 for the Square Root Correction. However, in just 11.48\% of
these conditions, the \(C_{llr}\) obtained using the Hapax Correction
was within 0.05 of the \(C_{llr}\) achieved by the baseline, and for the
Square Root Correction, this was the case in just 8.2\% of cases.
Comparing the proposed methods directly, the Hapax Correction performed
better than the Square Root Correction in 70\% of cases.
Figure~\ref{fig-avo} also reveals several anomalous results when
logistic regression calibration is implemented (left), as indicated
by multiple empty cells.

\subsubsection{\textit{Twitter}}\label{c_twitter}

\begin{landscape}
\begin{figure}

\centering{

\includegraphics[width=1\linewidth,height=\textheight,keepaspectratio, trim = 0 3cm 0 3cm, clip]{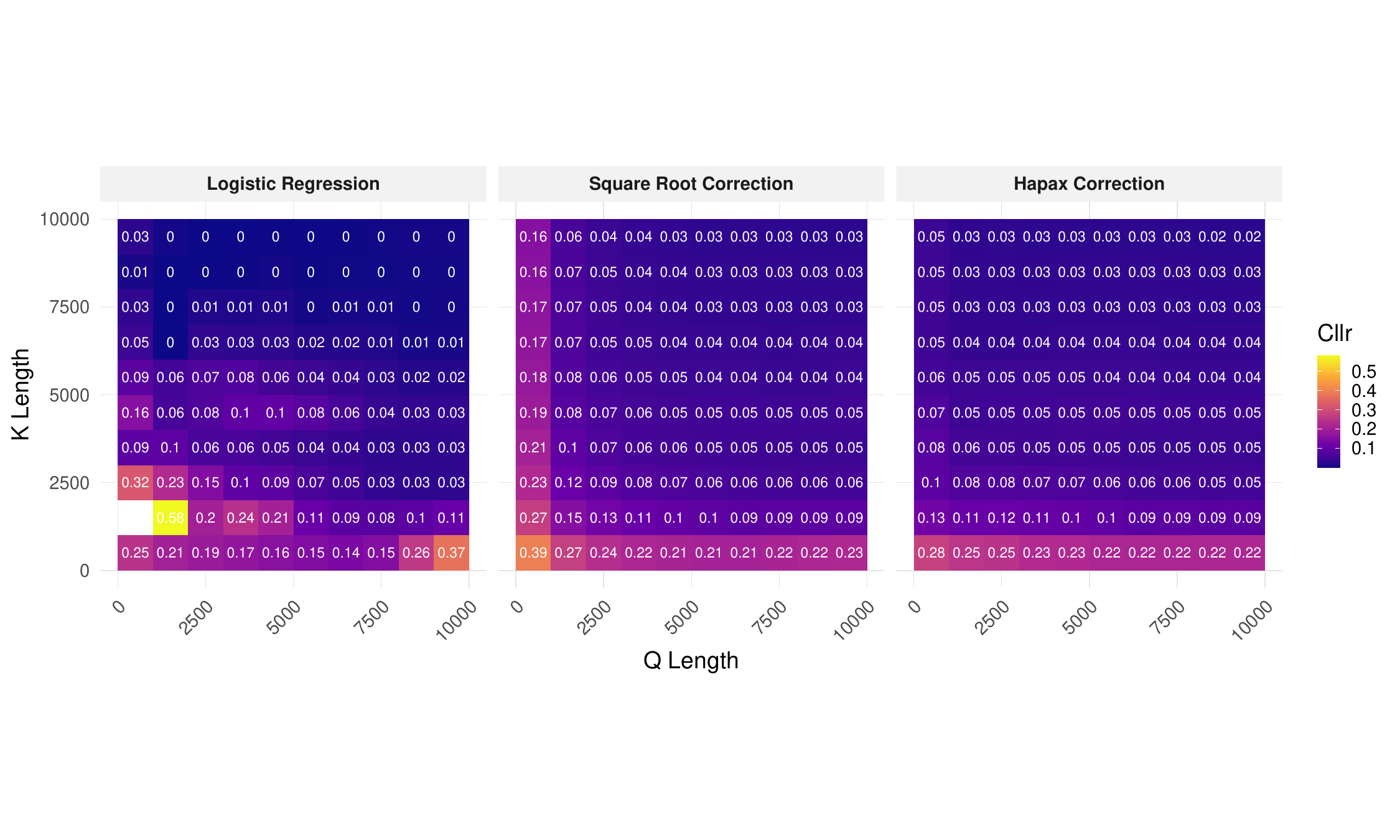}

}

\caption{\label{fig-twitter}Comparison of three approaches on the
Twitter dataset showing the relationship between Q Length, K Length and
mean Cllr. Each panel corresponds to a post-processing method applied to
LambdaG scores: logistic regression calibration (left), the Square Root
Correction (middle), and the Hapax Correction (right).}

\end{figure}
\end{landscape}

Figure~\ref{fig-twitter} shows the same analysis as
Figure~\ref{fig-avo}, applied to a register on which LambdaG has not yet
been formally tested: tweets. LambdaG demonstrates strong performance
when both logistic regression calibration and the proposed corrections
are implemented.

Figure~\ref{fig-twitter} shows the proposed approaches obtained
well-calibrated LLRs even when the case data was very short - \emph{K}
\(length = 500\) \(tokens\). This was the worst performing condition,
yet \(C_{llr}\) values remained consistently low (\(\le\) 0.39).
Contrastingly, for logistic regression calibration, when \emph{K} length
= 1,500, and \emph{Q} length \(\le\) 1,500, anomalously high \(C_{llr}\)
values are evidenced.

Across all the tested conditions, logistic regression calibration
achieved the best performance in 67\% of cases. However, the Hapax
Correction performs comparably overall. Where logistic regression
calibration obtained a \(C_{llr}\) \textgreater{} 0.1, the Hapax
Correction matched or outperformed it in 65.2\% of conditions, while the
Square Root Correction did so in 60.9\%. In conditions where logistic
regression calibration achieved \(C_{llr}\) \textless{} 0.1, the Hapax
Correction was always within 0.04, and the Square Root Correction was
within 0.05 in 87.01\% of cases. Directly comparing the Hapax Correction
and the Square Root Correction, the former outperforms the latter across
87\% of the tested conditions. Nevertheless, the performance of the
Square Root Correction remained within 5\% of the Hapax Correction
across 90\% of conditions.

\subsubsection{\textit{Bolt}}\label{c_bolt}

\begin{figure}

\centering{

\includegraphics[width=1\linewidth,keepaspectratio, trim = 0 2.5cm 0 2.5cm, clip]{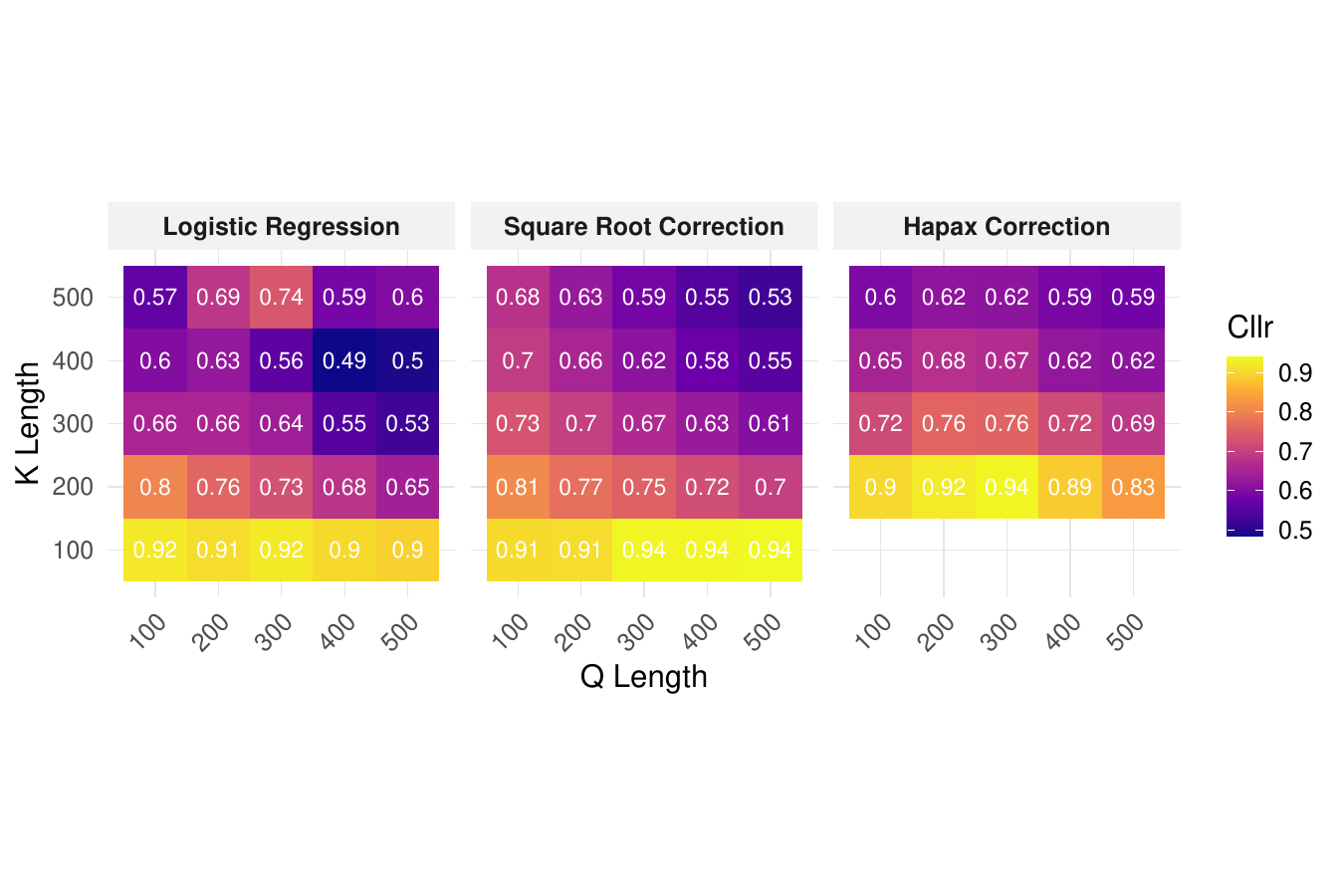}

}

\caption{\label{fig-bolt}Comparison of three approaches on the Bolt
dataset showing the relationship between Q Length, K Length and mean
Cllr. Each panel corresponds to a post-processing method applied to
LambdaG scores: logistic regression calibration (left), the Square Root Correction (middle), and the Hapax
Correction (right).}

\end{figure}

Figure~\ref{fig-bolt} illustrates the \(C_{llr}\) obtained when using
the three approaches to post-process \(\lambda_G\) calculated on the
text messages in \textit{Bolt}. The text lengths evaluated ranged from
100 to 500 tokens in increments of 100. Considering this very limited
amount of data, there is a strong performance across all three
approaches, particularly with a minimum of 200 tokens of \emph{Q} and
\emph{K} data. Performance tends to improve further as the amount of
data increases. Whilst logistic regression calibration outperforms both
the corrections across 80\% of conditions, in certain conditions the
Square Root Correction achieves the lowest \(C_{llr}\), for example when
\emph{Q} length = 500 and \emph{K} length = 500.

In contrast to the performance observed on the other registers, on this
dataset, the Square Root Correction tends to outperform the Hapax
Correction, exhibiting better performance across 84\% of the conditions
tested. Figure~\ref{fig-bolt} also indicates that, for this dataset, the
Hapax Correction requires more than 100 tokens of \emph{K} data to
achieve a meaningful output (\(C_{llr}\) below one).

\section{Discussion}\label{discussion}

This paper has compared two novel alternatives to the calibration of
\(\lambda_G\) against the conventional approach of logistic regression
calibration \citep{niniGrammarBehavioralBiometric2026}. The objective
was to eliminate the need to train inherently data-dependent,
time-consuming and opaque calibration models in order to produce simple,
accessible and well-calibrated measures of evidential strength for AV.
We believe both the Square Root Correction and the Hapax Correction
achieved this, with the Hapax Correction in particular demonstrating a
strong performance across the fifteen corpora used for evaluation.

Although the Hapax Correction achieved a lower win rate than the
baseline of logistic regression calibration, it still outperformed this
baseline in approximately 45\% of the tests, weighted by corpora.
Furthermore, in the cases where logistic regression calibration achieved
better performance than the Hapax Correction, the performance gap was
more frequently small in magnitude i.e.~less than 5\% relative
difference, than the inverse comparison. This highlights the consistency
in the performance of the Hapax Correction, as whilst logistic
regression calibration may achieve stronger peak performance, its
performance is also more variable.

The evaluation across varying \emph{Q} and \emph{K} lengths revealed an
interaction between input length and method performance. Logistic
regression calibration tends to outperform the Hapax Correction when
\emph{Q} and \emph{K} texts are longer, achieving near optimal
performance on the \textit{Avocado} and \textit{Twitter} when \emph{K} is
longer than 5,500 tokens. This may be due to the increased input
supporting more stable parameter estimation. Under these conditions, the
Hapax Correction also demonstrates strong performance, consistently
producing a \(C_{llr}\) below 0.1, indicating well-calibrated LLRs.
Comparatively, on shorter text lengths the Hapax Correction more
frequently outperforms logistic regression calibration, demonstrating
greater robustness with less linguistic input. This is of note because
these shorter text lengths are more likely to be practically relevant,
indicating that the Hapax Correction may offer more reliable performance
in real-world cases.

There were, however, certain conditions where the Hapax Correction did
not produce meaningful LLRs. This was demonstrated on \textit{Bolt} when
the \emph{K} sample was approximately 100 tokens. This is likely because
in a very short text the proportion of hapax legomena is extremely high.
This relates closely to Herdan-Heap's Law, which shows that vocabulary
growth is rapid at first before it slows down considerably
\citep{herdanTypetokenMathematics1960, heapsInformationRetrievalComputational1978a}.
This makes the measure noisy and unreliable as it is overestimating
uniqueness. Nonetheless, for most authorship analysis methods, a
\emph{K} sample of 100 tokens is likely insufficient
\citep{stamatatosSurveyModernAuthorship2009, stamatatosOverviewAuthorshipVerification2023}.
Both logistic regression calibration and the Square Root Correction also
obtained \(C_{llr}\) values greater than 0.9, indicating comparatively
poor calibration relative to the other conditions tested.

In other tests, logistic regression calibration obtained \(C_{llr}\)
values greater than one, indicating that the system is misleading
\citep{vanlieropOverviewLogLikelihood2024a}. The anomalous values were
seen in tests on \textit{Avocado} and \textit{Twitter}. These anomalies
are likely due to the small calibration datasets (\(40\) problems in
\textit{Avocado}; \(60\) in \textit{Twitter}), which makes the logistic
regression sensitive to noise and prone to overfitting. These results
highlight a key limitation of logistic regression calibration: its
reliance on substantial amounts of suitable calibration data, which is
often costly and difficult to obtain. In contrast, the proposed
corrections exhibit lower variability, suggesting greater robustness.

However, despite the greater robustness of the Square Root Correction,
its performance was not as competitive overall. It was outperformed by
both the Hapax Correction and logistic regression calibration across the
majority of tests, and where it was outperformed, it was typically not
within a 5\% margin of the other methods. That said, logistic regression calibration
represents a strong baseline, and the performance of the Hapax
Correction is particularly effective. The Square Root Correction is a
simple transformation, and despite this simplicity, it is still able to
outperform the logistic regression calibration baseline on certain
corpora. It is therefore still a meaningful and competitive approach.

Furthermore, both the Square Root correction and the Hapax Correction
move towards a more interpretable approach compared to logistic
regression calibration. By further aligning the assumptions underlying
LambdaG with our theoretical and probabilistic understanding of language
productivity, the proposed corrections move forensic AV towards the
generative modeling approach seen in state-of-the-art forensic sciences
such as DNA profiling \citep{baldingDNAProfileMatch1994a, gillReviewProbabilisticGenotyping2021, mitchellGenomicCodeGenome2025}. In these approaches, extraneous sources of
variability are incorporated directly into the likelihood function
\citep{taylorFactorsAffectingPeak2016}. This enables the true probability of the observed DNA evidence to be calculated under each competing hypothesis. As a result, LLRs are
computed directly from the known generative distributions \((P(x|y))\)
\citep{collinsLikelihoodRatiosDNA1994, buckletonProbabilisticGenotypingSoftware2019},
rather than derived through separate calibration procedures.

We hypothesised that repetition leads to artificially inflated
\(\lambda_G\) values, given the implicit assumption of conditional
independence in naïve-Bayes-based algorithms. The usage-based theories
underlying LambdaG suggest that individuals exhibit consistent
preferences for certain sequences
\citep{niniTheoryLinguisticIndividuality2023a}. This enables AV, as
individuals use the same sequences across different texts. However, this
also suggests that repetitions within a single text may not be
conditionally independent. Therefore, subsequent repetitions of the same
pattern should plausibly contribute less incremental evidential weight
than their initial occurrence.

This effect is consistent with established observations in lexical
statistics, such as the sub-linear growth of vocabulary described by
Herdan-Heap's law
\citep{herdanTypetokenMathematics1960, heapsInformationRetrievalComputational1978a}.
This law formalises the decrease in the rate of introduction of new
types as text length increases.

As a result of this increased redundancy, tokens appearing later in a
text are less likely to provide as much new stylistic evidence than
earlier tokens. Therefore, whilst \(\lambda_G\) grows approximately
linearly with the number of tokens in the disputed document, the amount
of effective new evidence grows sub-linearly. This creates a systematic
bias, where longer texts are overstated as disproportionately
informative under the raw additive score. This effect is believed to be
mitigated by the Square Root Correction.

When implementing the Hapax Correction, \(\lambda_G\) is multiplied by
the number of hapax legomena in the \emph{Q} text, divided by the total
number of tokens. This constitutes an existing statistical measure of
linguistic productivity called Baayen's \(\mathcal{P}\)
\citep[p.50]{baayenWordFrequencyDistributions2001b}. This metric can be
used to estimate the slope of Herdan-Heap's law. The Hapax correction could
therefore more precisely model the relationship between text
length, lexical productivity and inflated \(\lambda_G\) values.

Ultimately, the simplified approach achieved by both normalisations is important not only for experts
but to ensure accessibility for fact-finders. The proposed corrections
are more interpretable and transparent, which in turn facilitates more
reliable decision-making.

\section{Limitations}

The primary limitation of this research is the absence of formal mathematical proof specifying why the proposed corrections achieve such strong performance. The corrections have some theoretical grounding, as we hypothesis that they  proportionally scale $\lambda_G$ values that may be over-inflated due to repetition in the data. However, the reason underlying the impressive empirical performance of these specific metrics, as opposed to other measures of lexical diversity, is not yet fully understood. 

Furthering our understanding of these corrections is an avenue for future research and may enhance understanding of individual language productivity and the robustness of AV methods. Furthermore, It is plausible that both the proposed corrections are approximating the same underlying concept, which, if identifiable, could constitute the optimal approach. 

It is important to note that, as with logistic regression calibration, for each AV case the forensic linguist is required to validate the application of the proposed corrections to that context. As such, this partial understanding of the proposed corrections does not invalidate their use. 

A second limitation of this study is that the proposed methods have only been validated on English. As such, another direction for future research is evaluating the proposed corrections on texts in different languages to obtain further insight as to the robustness and underlying assumptions of the normalisations.

\section{Conclusion}\label{conclusion}

This paper demonstrates that by applying simple transformations to
LambdaG scores \citep{niniGrammarBehavioralBiometric2026}, it is
possible to generate meaningful LLRs without training a separate
calibration model. The transformations consist of dividing the LambdaG
score by the square root of the number of tokens in the disputed text,
or multiplying it by the ratio of hapax legomena to tokens in the text.

One of the primary limitations of training a separate calibration model
is the high dependency on data, which is often difficult and time
consuming to obtain and prepare. This process also relies on subjective
expert decisions regarding data selection and structure
(e.g.~anchoring). As there is currently limited justification in the
literature, these decisions typically require validation for each case.
This further increases the complexity of the method and the difficulty of
explaining results to non-experts. By removing the need for a separate
calibration model, the proposed corrections also eliminate these
limitations, as well as issues such as over- or underfitting, and the
inherently opaque nature of the calibration models.

The proposed corrections applied to LambdaG constitute what is currently
the only method of AV that can generate well-calibrated LLRs without a
separate calibration model. The corrections achieve this whilst
maintaining the state-of-the-art performance of LambdaG when calibrated
using logistic regression. This was tested across fifteen different
corpora, and text lengths ranging from 100 to 9,500 tokens. Overall, on
average across datasets, the Hapax Correction outperforms logistic
regression calibration approximately 45\% the time, with comparable
performance across the other instances. The performance of the Square
Root Correction was comparatively modest; however, its simplicity makes
the results noteworthy.

Furthermore, the proposed corrections are consistent with usage-based
linguistic theory and well-established lexical statistics. Although the
reasoning for their effectiveness is not yet fully understood, the
normalisations may account for the notion that repeated sequences within
a text provide diminishing evidential value with each occurrence. This
aligns with the principles of language productivity on which LambdaG is
based and offers greater scientific validity than logistic regression
calibration.

Overall, the proposed methods, particularly the Hapax Correction,
achieved calibration performance comparable to logistic regression
calibration across text lengths and registers. The combination of
empirical performance and unique practical advantages identified in this
paper supports the adoption of the proposed approaches in forensic
settings on English-language texts.

\renewcommand\refname{References}
\bibliography{references.bib}

@article{argamonInterpretingBurrowssDelta2008,
  title = {Interpreting {{Burrows}}'s {{Delta}}: {{Geometric}} and {{Probabilistic Foundations}}},
  shorttitle = {Interpreting {{Burrows}}'s {{Delta}}},
  author = {Argamon, Shlomo},
  year = 2008,
  month = jun,
  journal = {Literary and Linguistic Computing},
  volume = {23},
  number = {2},
  pages = {131--147},
  issn = {0268-1145},
  doi = {10.1093/llc/fqn003},
  urldate = {2026-04-08}
}

@book{baayenWordFrequencyDistributions2001b,
  title = {Word {{Frequency Distributions}}},
  author = {Baayen, R. Harald},
  year = 2001,
  series = {Text, {{Speech}} and {{Language Technology}}},
  publisher = {Springer Netherlands},
  address = {Dordrecht},
  doi = {10.1007/978-94-010-0844-0},
  urldate = {2026-01-14},
  langid = {english}
}

@article{baldingDNAProfileMatch1994a,
  title = {{{DNA}} Profile Match Probability Calculation: How to Allow for Population Stratification, Relatedness, Database Selection and Single Bands},
  shorttitle = {{{DNA}} Profile Match Probability Calculation},
  author = {Balding, David J. and Nichols, Richard A.},
  year = 1994,
  month = feb,
  journal = {Forensic Science International},
  volume = {64},
  number = {2},
  pages = {125--140},
  issn = {0379-0738},
  doi = {10.1016/0379-0738(94)90222-4},
  urldate = {2026-01-13}
}

@misc{benoitSpacyrWrapperSpaCy2023a,
  title = {Spacyr: {{Wrapper}} to the '{{spaCy}}' '{{NLP}}' {{Library}}},
  shorttitle = {Spacyr},
  author = {Benoit, Kenneth and Matsuo, Akitaka and Gruber, Johannes},
  year = 2023,
  month = dec,
  doi = {10.32614/CRAN.package.spacyr},
  copyright = {GPL-3}
}

@article{biberRegisterPredictorLinguistic2012,
  title = {Register as a Predictor of Linguistic Variation},
  author = {Biber, Douglas},
  year = 2012,
  month = may,
  doi = {10.1515/cllt-2012-0002},
  urldate = {2026-04-10},
  chapter = {Corpus Linguistics and Linguistic Theory},
  copyright = {De Gruyter expressly reserves the right to use all content for commercial text and data mining within the meaning of Section 44b of the German Copyright Act.},
  langid = {english}
}

@article{biedermannReframingDebateQuestion2016,
  title = {Reframing the Debate: {{A}} Question of Probability, Not of Likelihood Ratio},
  shorttitle = {Reframing the Debate},
  author = {Biedermann, Alex and Bozza, Silvia and Taroni, Franco T. and Aitken, Colin Graeme Girdwood},
  year = 2016,
  month = sep,
  journal = {Science \& Justice},
  volume = {56},
  number = {5},
  pages = {392--396},
  issn = {1355-0306},
  doi = {10.1016/j.scijus.2016.05.008},
  urldate = {2026-01-13}
}

@article{biosaEvaluationForensicData2020,
  title = {Evaluation of {{Forensic Data Using Logistic Regression-Based Classification Methods}} and an {{R Shiny Implementation}}},
  author = {Biosa, Giulia and Giurghita, Diana and Vincenti, Marco and Neocleous, Tereza},
  year = 2020,
  month = oct,
  journal = {Frontiers in Chemistry},
  volume = {8},
  pages = {738},
  issn = {2296-2646},
  doi = {10.3389/fchem.2020.00738},
  urldate = {2026-04-08},
  pmcid = {PMC7609892},
  pmid = {33195014}
}

@inproceedings{boenninghoffSimilarityLearningAuthorship2019,
  title = {Similarity {{Learning}} for {{Authorship Verification}} in {{Social Media}}},
  booktitle = {International {{Conference}} on {{Acoustics}}, {{Speech}} and {{Signal Processing}}},
  author = {Boenninghoff, Benedikt and Nickel, Robert and Zeiler, Steffen and Kolossa, Dorothea},
  year = 2019,
  doi = {10.1109/ICASSP.2019.8683405},
  langid = {english}
}

@article{brummerApplicationindependentEvaluationSpeaker2006,
  title = {Application-Independent Evaluation of Speaker Detection},
  author = {Br{\"u}mmer, Niko and {du Preez}, Johan},
  year = 2006,
  month = apr,
  journal = {Computer Speech \& Language},
  series = {Odyssey 2004: {{The}} Speaker and {{Language Recognition Workshop}}},
  volume = {20},
  number = {2},
  pages = {230--275},
  issn = {0885-2308},
  doi = {10.1016/j.csl.2005.08.001},
  urldate = {2026-01-13}
}

@inproceedings{brummerLikelihoodratioCalibrationUsing2013,
  title = {Likelihood-Ratio Calibration Using Prior-Weighted Proper Scoring Rules},
  booktitle = {Interspeech 2013},
  author = {Br{\"u}mmer, Niko and Doddington, George R.},
  year = 2013,
  month = aug,
  publisher = {ISCA},
  doi = {10.21437/Interspeech.2013-470},
  urldate = {2026-04-08},
  langid = {english}
}

@article{buckletonProbabilisticGenotypingSoftware2019,
  title = {The {{Probabilistic Genotyping Software STRmix}}: {{Utility}} and {{Evidence}} for Its {{Validity}}},
  shorttitle = {The {{Probabilistic Genotyping Software STRmix}}},
  author = {Buckleton, John S. and Bright, Jo-Anne and Gittelson, Simone and Moretti, Tamyra R. and Onorato, Anthony J. and Bieber, Frederick R. and Budowle, Bruce and Taylor, Duncan A.},
  year = 2019,
  journal = {Journal of Forensic Sciences},
  volume = {64},
  number = {2},
  pages = {393--405},
  issn = {1556-4029},
  doi = {10.1111/1556-4029.13898},
  urldate = {2026-03-04},
  copyright = {\copyright{} 2018 American Academy of Forensic Sciences},
  langid = {english}
}

@article{chenEmpiricalStudySmoothing1999,
  title = {An Empirical Study of Smoothing Techniques for Language Modeling},
  author = {Chen, Stanley F. and Goodman, Joshua},
  year = 1999,
  month = oct,
  journal = {Computer Speech \& Language},
  volume = {13},
  number = {4},
  pages = {359--394},
  issn = {0885-2308},
  doi = {10.1006/csla.1999.0128}
}

@misc{chensongBOLTEnglishSMS2018,
  title = {{{BOLT English SMS}}/{{Chat}}},
  author = {{Chen, Song} and {Fore, Dana} and {Strassel, Stephanie} and {Lee, Haejoong} and {Wright, Jonathan}},
  year = 2018,
  month = aug,
  pages = {179792 KB},
  publisher = {Linguistic Data Consortium},
  doi = {10.35111/HKFC-7865}
}

@article{collinsLikelihoodRatiosDNA1994,
  title = {Likelihood Ratios for {{DNA}} Identification},
  author = {Collins, Andrew and Morton, Newton E.},
  year = 1994,
  month = jul,
  journal = {Proceedings of the National Academy of Sciences of the United States of America},
  volume = {91},
  pages = {6007--11},
  doi = {10.1073/pnas.91.13.6007}
}

@article{coulthardAuthorIdentificationIdiolect2004,
  title = {Author {{Identification}}, {{Idiolect}}, and {{Linguistic Uniqueness}}},
  author = {Coulthard, Malcolm},
  year = 2004,
  month = dec,
  journal = {Applied Linguistics},
  volume = {25},
  number = {4},
  pages = {431--447},
  issn = {0142-6001},
  doi = {10.1093/applin/25.4.431},
  urldate = {2026-01-14}
}

@article{dawidWellCalibratedBayesian1982,
  title = {The {{Well-Calibrated Bayesian}}},
  author = {Dawid, A. P.},
  year = 1982,
  month = sep,
  journal = {Journal of the American Statistical Association},
  volume = {77},
  number = {379},
  pages = {605--610},
  publisher = {Taylor \& Francis},
  issn = {0162-1459},
  doi = {10.1080/01621459.1982.10477856},
  urldate = {2026-01-13}
}

@article{degrootComparisonEvaluationForecasters1983,
  title = {The {{Comparison}} and {{Evaluation}} of {{Forecasters}}},
  author = {DeGroot, Morris H. and Fienberg, Stephen E.},
  year = 1983,
  journal = {Journal of the Royal Statistical Society. Series D (The Statistician)},
  volume = {32},
  number = {1},
  eprint = {2987588},
  eprinttype = {jstor},
  pages = {12--22},
  publisher = {[Royal Statistical Society, Wiley]},
  issn = {0039-0526},
  doi = {10.2307/2987588},
  urldate = {2026-01-13}
}

@article{evertUnderstandingExplainingDelta2017,
  title = {Understanding and Explaining {{Delta}} Measures for Authorship Attribution},
  author = {Evert, Stefan and Proisl, Thomas and Jannidis, Fotis and Reger, Isabella and Pielstr{\"o}m, Steffen and Sch{\"o}ch, Christof and Vitt, Thorsten},
  year = 2017,
  month = dec,
  journal = {Digital Scholarship in the Humanities},
  volume = {32},
  number = {2},
  pages = {ii4-ii16},
  issn = {2055-7671},
  doi = {10.1093/llc/fqx023},
  urldate = {2026-03-02}
}

@techreport{forensicscienceregulatorCodesPracticeConduct2021,
  title = {Codes of {{Practice}} and {{Conduct}}},
  author = {{Forensic Science Regulator}},
  year = 2021,
  number = {FSR-C-118},
  address = {United Kingdom}
}

@article{gillReviewProbabilisticGenotyping2021,
  title = {A {{Review}} of {{Probabilistic Genotyping Systems}}: {{EuroForMix}}, {{DNAStatistX}} and {{STRmix}}™},
  shorttitle = {A {{Review}} of {{Probabilistic Genotyping Systems}}},
  author = {Gill, Peter and Benschop, Corina and Buckleton, John and Bleka, {\O}yvind and Taylor, Duncan},
  year = 2021,
  month = sep,
  journal = {Genes},
  volume = {12},
  number = {10},
  pages = {1559},
  issn = {2073-4425},
  doi = {10.3390/genes12101559},
  urldate = {2026-03-04},
  pmcid = {PMC8535381},
  pmid = {34680954}
}

@article{gonzalez-rodriguezEmulatingDNARigorous2007,
  title = {Emulating {{DNA}}: {{Rigorous Quantification}} of {{Evidential Weight}} in {{Transparent}} and {{Testable Forensic Speaker Recognition}}},
  shorttitle = {Emulating {{DNA}}},
  author = {{Gonzalez-Rodriguez}, Joaquin and Rose, Phil and Ramos, Daniel and Toledano, Doroteo T. and {Ortega-Garcia}, Javier},
  year = 2007,
  month = sep,
  journal = {IEEE Transactions on Audio, Speech, and Language Processing},
  volume = {15},
  number = {7},
  pages = {2104--2115},
  issn = {1558-7924},
  doi = {10.1109/TASL.2007.902747},
  urldate = {2026-01-13}
}

@book{grantIdeaProgressForensic2022a,
  title = {The {{Idea}} of {{Progress}} in {{Forensic Authorship Analysis}}},
  author = {Grant, Tim},
  year = 2022,
  series = {Elements in {{Forensic Linguistics}}},
  publisher = {Cambridge University Press},
  address = {Cambridge},
  doi = {10.1017/9781108974714},
  urldate = {2026-01-13},
  isbn = {978-1-108-97132-4}
}

@article{grieveAttributingBixbyLetter2019a,
  title = {Attributing the {{Bixby Letter}} Using N-Gram Tracing},
  author = {Grieve, Jack and Clarke, Isobelle and Chiang, Emily and Gideon, Hannah and Heini, Annina and Nini, Andrea and Waibel, Emily},
  year = 2019,
  month = sep,
  journal = {Digital Scholarship in the Humanities},
  volume = {34},
  number = {3},
  pages = {493--512},
  issn = {2055-7671, 2055-768X},
  doi = {10.1093/llc/fqy042},
  urldate = {2026-03-02},
  copyright = {https://academic.oup.com/journals/pages/open\_access/funder\_policies/chorus/standard\_publication\_model},
  langid = {english}
}

@article{grieveFrontiersMappingLexical2019,
  title = {Frontiers \textbar{} {{Mapping Lexical Dialect Variation}} in {{British English Using Twitter}}},
  author = {Grieve, Jack and Montgomery, Chris and Nini, Andrea and Murakami, Akira and Guo, Diansheng},
  year = 2019,
  doi = {10.3389/frai.2019.00011},
  langid = {english}
}

@article{guoCalibrationModernNeural2017a,
  title = {On {{Calibration}} of {{Modern Neural Networks}}},
  author = {Guo, Chuan and Pleiss, Geoff and Sun, Yu and Weinberger, Kilian Q.},
  year = 2017,
  month = jun,
  journal = {ArXiv},
  doi = {10.48550/arXiv.1706.04599}
}

@inproceedings{halvaniPOSNoiseEffectiveCountermeasure2021a,
  title = {{{POSNoise}}: {{An Effective Countermeasure Against Topic Biases}} in {{Authorship Analysis}}},
  shorttitle = {{{POSNoise}}},
  booktitle = {Proceedings of the 16th {{International Conference}} on {{Availability}}, {{Reliability}} and {{Security}}},
  author = {Halvani, Oren and Graner, Lukas},
  year = 2021,
  month = aug,
  series = {{{ARES}} '21},
  pages = {1--12},
  publisher = {Association for Computing Machinery},
  address = {New York, NY, USA},
  doi = {10.1145/3465481.3470050},
  isbn = {978-1-4503-9051-4}
}

@inproceedings{halvaniUsefulnessCompressionModels2017,
  title = {On the {{Usefulness}} of {{Compression Models}} for {{Authorship Verification}}},
  booktitle = {Proceedings of the 12th {{International Conference}} on {{Availability}}, {{Reliability}} and {{Security}}},
  author = {Halvani, Oren and Winter, Christian and Graner, Lukas},
  year = 2017,
  month = aug,
  series = {{{ARES}} '17},
  pages = {1--10},
  publisher = {Association for Computing Machinery},
  address = {New York, NY, USA},
  doi = {10.1145/3098954.3104050},
  urldate = {2026-04-16},
  isbn = {978-1-4503-5257-4}
}

@book{heapsInformationRetrievalComputational1978a,
  title = {Information {{Retrieval}}, {{Computational}} and {{Theoretical Aspects}}},
  author = {Heaps, H. S.},
  year = 1978,
  publisher = {Academic Press},
  googlebooks = {JWiqqBYbB0sC},
  isbn = {978-0-12-335750-2},
  langid = {english}
}

@article{heplerScorebasedLikelihoodRatios2012,
  title = {Score-Based Likelihood Ratios for Handwriting Evidence},
  author = {Hepler, Amanda B. and Saunders, Christopher P. and Davis, Linda J. and Buscaglia, JoAnn},
  year = 2012,
  month = jun,
  journal = {Forensic Science International},
  volume = {219},
  number = {1},
  pages = {129--140},
  issn = {0379-0738},
  doi = {10.1016/j.forsciint.2011.12.009}
}

@book{herdanTypetokenMathematics1960,
  title = {Type-Token {{Mathematics}}},
  author = {Herdan, Gustav},
  year = 1960,
  publisher = {Mouton},
  googlebooks = {LQtZAAAAMAAJ},
  langid = {english}
}

@article{huertas-tatoUnderstandingWritingStyle2024,
  title = {Understanding Writing Style in Social Media with a Supervised Contrastively Pre-Trained Transformer},
  author = {{Huertas-Tato}, Javier and Mart{\'i}n, Alejandro and Camacho, David},
  year = 2024,
  month = jul,
  journal = {Knowledge-Based Systems},
  volume = {296},
  pages = {111867},
  issn = {0950-7051},
  doi = {10.1016/j.knosys.2024.111867}
}

@inproceedings{ishiharaEstimatingStrengthAuthorship2022,
  title = {Estimating the {{Strength}} of {{Authorship Evidence}} with a {{Deep-Learning-Based Approach}}},
  booktitle = {Proceedings of the 20th {{Annual Workshop}} of the {{Australasian Language Technology Association}}},
  author = {Ishihara, Shunichi and Tsuge, Satoru and Inaba, Mitsuyuki and Zaitsu, Wataru},
  editor = {Parameswaran, Pradeesh and Biggs, Jennifer and Powers, David},
  year = 2022,
  month = dec,
  pages = {183--187},
  publisher = {Australasian Language Technology Association},
  address = {Adelaide, Australia}
}

@inproceedings{ishiharaForensicAuthorshipClassification2011,
  title = {A {{Forensic Authorship Classification}} in {{SMS Messages}}: {{A Likelihood Ratio Based Approach Using N-gram}}},
  shorttitle = {A {{Forensic Authorship Classification}} in {{SMS Messages}}},
  booktitle = {Proceedings of the {{Australasian Language Technology Association Workshop}} 2011},
  author = {Ishihara, Shunichi},
  editor = {Molla, Diego and Martinez, David},
  year = 2011,
  month = dec,
  pages = {47--56},
  address = {Canberra, Australia}
}

@article{ishiharaScorebasedLikelihoodRatios2021a,
  title = {Score-Based Likelihood Ratios for Linguistic Text Evidence with a Bag-of-Words Model},
  author = {Ishihara, Shunichi},
  year = 2021,
  month = oct,
  journal = {Forensic Science International},
  volume = {327},
  issn = {0379-0738},
  doi = {10.1016/j.forsciint.2021.110980},
  urldate = {2026-01-13}
}

@article{ishiharaStrengthForensicText2017,
  title = {Strength of Forensic Text Comparison Evidence from Stylometric Features: {{A}} Multivariate Likelihood Ratio-Based Analysis},
  shorttitle = {Strength of Forensic Text Comparison Evidence from Stylometric Features},
  author = {Ishihara, Shunichi},
  year = 2017,
  month = may,
  journal = {International Journal of Speech Language and the Law},
  volume = {24},
  pages = {67--98},
  doi = {10.1558/ijsll.30305}
}

@article{juolaVerifyingAuthorshipForensic2021b,
  title = {Verifying Authorship for Forensic Purposes: {{A}} Computational Protocol and Its Validation},
  shorttitle = {Verifying Authorship for Forensic Purposes},
  author = {Juola, Patrick},
  year = 2021,
  month = aug,
  journal = {Forensic Science International},
  volume = {325},
  issn = {0379-0738},
  doi = {10.1016/j.forsciint.2021.110824},
  urldate = {2026-01-13}
}

@book{jurafskySpeechLanguageProcessing2026,
  title = {Speech and {{Language Processing}}: {{An Introduction}} to {{Natural Language Processing}}, {{Computational Linguistics}}, and {{Speech Recognition}}, with {{Language Models}}},
  author = {Jurafsky, Daniel and Martin, James H.},
  year = 2026,
  edition = {3}
}

@inproceedings{kneserImprovedBackingoffMgram1995,
  title = {Improved Backing-off for {{M-gram}} Language Modeling},
  booktitle = {1995 {{International Conference}} on {{Acoustics}}, {{Speech}}, and {{Signal Processing}}},
  author = {Kneser, {\relax Reinnhard}. and Ney, H.},
  year = 1995,
  month = may,
  volume = {1},
  pages = {181-184 vol.1},
  issn = {1520-6149},
  doi = {10.1109/ICASSP.1995.479394},
  urldate = {2026-01-20}
}

@article{koppelDeterminingIfTwo2014a,
  title = {Determining If Two Documents Are Written by the Same Author},
  author = {Koppel, Moshe and Winter, Yaron},
  year = 2014,
  journal = {Journal of the Association for Information Science and Technology},
  volume = {65},
  number = {1},
  pages = {178--187},
  issn = {2330-1643},
  doi = {10.1002/asi.22954},
  copyright = {\copyright{} 2013 ASIS\&T},
  langid = {english}
}

@article{koppelFundamentalProblemAuthorship2012a,
  title = {The ``{{Fundamental Problem}}'' of {{Authorship Attribution}}},
  author = {Koppel, Moshe and Schler, Jonathan and Argamon, Shlomo and Winter, Yaron},
  year = 2012,
  month = may,
  journal = {English Studies},
  volume = {93},
  number = {3},
  pages = {284--291},
  publisher = {Routledge},
  issn = {0013-838X},
  doi = {10.1080/0013838X.2012.668794},
  urldate = {2026-01-13}
}

@article{macarullarodriguezImprovedLikelihoodRatios2024,
  title = {Improved Likelihood Ratios for Face Recognition in Surveillance Video by Multimodal Feature Pairing},
  author = {Macarulla Rodriguez, Andrea and Geradts, Zeno and Worring, Marcel and Unzueta, Luis},
  year = 2024,
  month = jan,
  journal = {Forensic Science International: Synergy},
  volume = {8},
  pages = {100458},
  issn = {2589-871X},
  doi = {10.1016/j.fsisyn.2024.100458},
  urldate = {2026-04-08}
}

@article{mitchellGenomicCodeGenome2025,
  title = {The {{Genomic Code}}: The Genome Instantiates a Generative Model of the Organism},
  shorttitle = {The {{Genomic Code}}},
  author = {Mitchell, Kevin J. and Cheney, Nick},
  year = 2025,
  month = jun,
  journal = {Trends in Genetics},
  volume = {41},
  number = {6},
  pages = {462--479},
  issn = {0168-9525},
  doi = {10.1016/j.tig.2025.01.008},
  urldate = {2026-03-04}
}

@article{mollinEntirelyUnderstandBlairism2009,
  title = {``{{I}} Entirely Understand'' Is a {{Blairism}}: {{The}} Methodology of Identifying Idiolectal Collocations},
  shorttitle = {``{{I}} Entirely Understand'' Is a {{Blairism}}},
  author = {Mollin, Sandra},
  year = 2009,
  month = aug,
  publisher = {John Benjamins},
  doi = {10.1075/ijcl.14.3.04mol},
  langid = {english}
}

@article{morrisonBiGaussianizedCalibrationLikelihood2024a,
  title = {Bi-{{Gaussianized}} Calibration of Likelihood Ratios},
  author = {Morrison, Geoffrey Stewart},
  year = 2024,
  month = jan,
  journal = {Law, Probability and Risk},
  volume = {23},
  number = {1},
  issn = {1470-8396},
  doi = {10.1093/lpr/mgae004},
  urldate = {2026-01-13}
}

@article{morrisonContextForensicCasework2021,
  title = {In the Context of Forensic Casework, Are There Meaningful Metrics of the Degree of Calibration?},
  author = {Morrison, Geoffrey Stewart},
  year = 2021,
  month = jan,
  journal = {Forensic Science International: Synergy},
  volume = {3},
  issn = {2589-871X},
  doi = {10.1016/j.fsisyn.2021.100157},
  urldate = {2026-01-13}
}

@article{morrisonEmpiricalEstimatePrecision2011,
  title = {An Empirical Estimate of the Precision of Likelihood Ratios from a Forensic-Voice-Comparison System},
  author = {Morrison, Geoffrey Stewart and Zhang, Cuiling and Rose, Philip},
  year = 2011,
  month = may,
  journal = {Forensic Science International},
  volume = {208},
  number = {1},
  pages = {59--65},
  issn = {0379-0738},
  doi = {10.1016/j.forsciint.2010.11.001},
  urldate = {2026-01-13}
}

@article{morrisonMeasuringValidityReliability2011a,
  title = {Measuring the Validity and Reliability of Forensic Likelihood-Ratio Systems},
  author = {Morrison, Geoffrey Stewart},
  year = 2011,
  month = sep,
  journal = {Science \& Justice},
  volume = {51},
  number = {3},
  pages = {91--98},
  issn = {1355-0306},
  doi = {10.1016/j.scijus.2011.03.002},
  urldate = {2026-01-13}
}

@article{reindersSourceanchoredTraceanchoredGeneral2022,
  title = {Source-Anchored, Trace-Anchored, and General Match Score-Based Likelihood Ratios for Camera Device Identification},
  author = {Reinders, Stephanie and Guan, Yong and Ommen, Danica and Newman, Jennifer},
  year = 2022,
  journal = {Journal of Forensic Sciences},
  volume = {67},
  number = {3},
  pages = {975--988},
  issn = {1556-4029},
  doi = {10.1111/1556-4029.14991},
  urldate = {2026-07-02},
  copyright = {\copyright{} 2022 The Authors. Journal of Forensic Sciences published by Wiley Periodicals LLC on behalf of American Academy of Forensic Sciences.},
  langid = {english}
}

@misc{ForensicAuthorshipAnalysis,
  title = {Forensic {{Authorship Analysis}}},
  author = {Roemling, Dana and Grieve, Jack},
  year = 2024,
  month = mar,
  journal = {Centre For Research and Evidence on Security Threats},
  urldate = {2026-07-02},
  howpublished = {https://crestresearch.ac.uk/comment/forensic-authorship-analysis/},
  langid = {british}
}

@incollection{grantStarbuckCase2022,
  title = {The {{Starbuck Case}}},
  booktitle = {Methodologies and {{Challenges}} in {{Forensic Linguistic Casework}}},
  author = {Grant, Tim and Grieve, Jack},
  year = 2022,
  pages = {13--28},
  publisher = {John Wiley \& Sons, Ltd},
  doi = {10.1002/9781394266661.ch2},
  urldate = {2026-07-02},
  chapter = {2},
  isbn = {978-1-394-26666-1},
  langid = {english}
}

@inproceedings{rivera-sotoLearningUniversalAuthorship2021,
  title = {Learning {{Universal Authorship Representations}}},
  booktitle = {Proceedings of the 2021 {{Conference}} on {{Empirical Methods}} in {{Natural Language Processing}}},
  author = {{Rivera-Soto}, Rafael A. and Miano, Olivia Elizabeth and Ordonez, Juanita and Chen, Barry Y. and Khan, Aleem and Bishop, Marcus and Andrews, Nicholas},
  editor = {Moens, Marie-Francine and Huang, Xuanjing and Specia, Lucia and Yih, Scott Wen-tau},
  year = 2021,
  month = nov,
  pages = {913--919},
  publisher = {Association for Computational Linguistics},
  address = {Online and Punta Cana, Dominican Republic},
  doi = {10.18653/v1/2021.emnlp-main.70},
  urldate = {2026-07-02}
}

@article{niniIdiolectPackageForensic,
  title = {Idiolect: {{An R}} Package for Forensic Authorship Analysis},
  shorttitle = {Idiolect},
  author = {Nini, Andrea},
  year = 2026,
  volume = {11},
  number = {119},
  publisher = {Journal of Open Source Software},
  urldate = {2026-07-02},
  langid = {english}
}

@article{morrisonTutorialLogisticregressionCalibration2013,
  title = {Tutorial on Logistic-Regression Calibration and Fusion:Converting a Score to a Likelihood Ratio},
  shorttitle = {Tutorial on Logistic-Regression Calibration and Fusion},
  author = {Morrison, Geoffrey Stewart},
  year = 2013,
  month = jun,
  journal = {Australian Journal of Forensic Sciences},
  volume = {45},
  number = {2},
  pages = {173--197},
  publisher = {Taylor \& Francis},
  issn = {0045-0618},
  doi = {10.1080/00450618.2012.733025},
  urldate = {2026-01-13}
}

@misc{mullenMinhashLocalitysensitiveHashing2020,
  title = {Minhash and Locality-Sensitive Hashing},
  author = {Mullen, Lincoln},
  year = 2020,
  urldate = {2026-01-23},
  howpublished = {https://cran.r-project.org/web/packages/textreuse/vignettes/textreuse-minhash.html}
}

@misc{niniIdiolectPackageForensic2024a,
  title = {Idiolect: {{An R}} Package for Forensic Authorship Analysis},
  author = {Nini, Andrea},
  year = 2024,
  doi = {10.32614/CRAN.package.idiolect},
  copyright = {GPL-2.0}
}

@article{niniRegisterVariationMalicious2017,
  title = {Register Variation in Malicious Forensic Texts},
  author = {Nini, Andrea},
  year = 2017,
  month = jun,
  journal = {The International Journal of Speech, Language and the Law},
  volume = {24},
  number = {1},
  pages = {99--126},
  publisher = {University of Toronto Press},
  issn = {1748-8885},
  doi = {10.1558/ijsll.30173},
  urldate = {2026-04-10}
}

@book{niniTheoryLinguisticIndividuality2023a,
  title = {A {{Theory}} of {{Linguistic Individuality}} for {{Authorship Analysis}}},
  author = {Nini, Andrea},
  year = 2023,
  series = {Elements in {{Forensic Linguistics}}},
  publisher = {Cambridge University Press},
  address = {Cambridge},
  doi = {10.1017/9781108974851},
  isbn = {978-1-108-97138-6}
}

@misc{oarddouglasAvocadoResearchEmail2015,
  title = {Avocado {{Research Email Collection}}},
  author = {{Oard, Douglas} and {Webber, William} and {Kirsch, David A.} and {Golitsynskiy, Sergey}},
  year = 2015,
  month = feb,
  pages = {4407648 KB},
  publisher = {Linguistic Data Consortium},
  doi = {10.35111/WQT6-JG60}
}

@inproceedings{pothaImprovedImpostorsMethod2017a,
  title = {An {{Improved Impostors Method}} for {{Authorship Verification}}},
  author = {Potha, Nektaria and Stamatatos, Efstathios},
  year = 2017,
  month = aug,
  doi = {10.1007/978-3-319-65813-1_14},
  urldate = {2026-01-20},
  langid = {english}
}

@inproceedings{pothaProfileBasedMethodAuthorship2014,
  title = {A {{Profile-Based Method}} for {{Authorship Verification}}},
  booktitle = {Artificial {{Intelligence}}:{{Methods}} and {{Application}}},
  author = {Potha, Nektaria and Stamatatos, Efstathios},
  year = 2014,
  series = {Lecture {{Notes}} in {{Computer Science}}},
  pages = {312--326},
  doi = {10.1007/978-3-319-07064-3_25},
  urldate = {2026-01-13},
  langid = {english}
}

@misc{pullBayesianApproachProbability2025,
  type = {{{SSRN Scholarly Paper}}},
  title = {A {{Bayesian Approach}} to {{Probability Default Model Calibration}}: {{Theoretical}} and {{Empirical Insights}} on the {{Jeffreys Test}}},
  shorttitle = {A {{Bayesian Approach}} to {{Probability Default Model Calibration}}},
  author = {Pull, Yoann and Hurlin, Christophe},
  year = 2025,
  number = {5291474},
  eprint = {5291474},
  publisher = {Social Science Research Network},
  address = {Rochester, NY},
  doi = {10.2139/ssrn.5291474},
  urldate = {2026-01-13},
  archiveprefix = {Social Science Research Network},
  langid = {english}
}

@article{ramosReliableSupportMeasuring2013,
  title = {Reliable Support: {{Measuring}} Calibration of Likelihood Ratios},
  shorttitle = {Reliable Support},
  author = {Ramos, Daniel and {Gonzalez-Rodriguez}, Joaquin},
  year = 2013,
  journal = {Forensic Science International},
  volume = {230},
  number = {1-3},
  pages = {156--69},
  doi = {10.1016/j.forsciint.2013.04.014},
  urldate = {2026-01-13},
  langid = {english}
}

@article{roseTechnicalForensicSpeaker2006a,
  title = {Technical Forensic Speaker Recognition: {{Evaluation}}, Types and Testing of Evidence},
  shorttitle = {Technical Forensic Speaker Recognition},
  author = {Rose, Phil},
  year = 2006,
  month = apr,
  journal = {Computer Speech \& Language},
  series = {Odyssey 2004: {{The}} Speaker and {{Language Recognition Workshop}}},
  volume = {20},
  number = {2},
  pages = {159--191},
  issn = {0885-2308},
  doi = {10.1016/j.csl.2005.07.003},
  urldate = {2026-01-13}
}

@article{silvafilhoClassifierCalibrationSurvey2023a,
  title = {Classifier Calibration: A Survey on How to Assess and Improve Predicted Class Probabilities},
  shorttitle = {Classifier Calibration},
  author = {Silva Filho, Telmo and Song, Hao and {Perello-Nieto}, Miquel and {Santos-Rodriguez}, Raul and Kull, Meelis and Flach, Peter},
  year = 2023,
  month = may,
  journal = {Machine Learning},
  volume = {112},
  number = {9},
  pages = {3211--3260},
  publisher = {Springer},
  issn = {1573-0565},
  doi = {10.1007/s10994-023-06336-7},
  urldate = {2026-01-13},
  copyright = {2023 The Author(s)},
  langid = {english}
}

@inproceedings{stamatatosOverviewAuthorshipVerification2023,
  title = {Overview of the {{Authorship Verification Task}} at {{PAN}} 2023},
  booktitle = {Conference and {{Labs}} of the {{Evaluation Forum}}},
  author = {Stamatatos, Efstathios and Kredens, Krzysztof and Pezik, Piotr and Heini, Annina and Bevendorff, Janek and Stein, Benno and Potthast, Martin},
  year = 2023
}

@article{stamatatosSurveyModernAuthorship2009,
  title = {A Survey of Modern Authorship Attribution Methods},
  author = {Stamatatos, Efstathios},
  year = 2009,
  journal = {Journal of the American Society for Information Science and Technology},
  volume = {60},
  number = {3},
  pages = {538--556},
  issn = {1532-2890},
  doi = {10.1002/asi.21001},
  urldate = {2026-04-08},
  copyright = {\copyright{} 2008 ASIS\&T},
  langid = {english}
}

@article{taylorFactorsAffectingPeak2016,
  title = {Factors Affecting Peak Height Variability for Short Tandem Repeat Data},
  author = {Taylor, Duncan and Buckleton, John and Bright, Jo-Anne},
  year = 2016,
  month = mar,
  journal = {Forensic Science International: Genetics},
  volume = {21},
  pages = {126--133},
  issn = {1872-4973},
  doi = {10.1016/j.fsigen.2015.12.009},
  urldate = {2026-03-04}
}

@article{taylorInterpretationSingleSource2013a,
  title = {The Interpretation of Single Source and Mixed {{DNA}} Profiles},
  author = {Taylor, Duncan and Bright, Jo-Anne and Buckleton, John},
  year = 2013,
  journal = {Forensic Science International. Genetics},
  volume = {7},
  number = {5},
  pages = {516--528},
  doi = {10.1016/j.fsigen.2013.05.011},
  urldate = {2026-01-13},
  langid = {english}
}

@inproceedings{vandervloedInterchangeabilityCalibrationAudio2024,
  title = {Interchangeability of {{Calibration Audio Datasets}} for {{Forensic Automatic Speaker Recognition}}},
  booktitle = {2024 12th {{International Workshop}} on {{Biometrics}} and {{Forensics}} ({{IWBF}})},
  author = {{van der Vloed}, David},
  year = 2024,
  month = apr,
  pages = {1--6},
  doi = {10.1109/IWBF62628.2024.10593938},
  urldate = {2026-01-13}
}

@article{vanlieropOverviewLogLikelihood2024a,
  title = {An Overview of Log Likelihood Ratio Cost in Forensic Science -- {{Where}} Is It Used and What Values Can We Expect?},
  author = {{van Lierop}, Stijn and Ramos, Daniel and Sjerps, Marjan and Ypma, Rolf},
  year = 2024,
  month = jan,
  journal = {Forensic Science International: Synergy},
  volume = {8},
  issn = {2589-871X},
  doi = {10.1016/j.fsisyn.2024.100466},
  urldate = {2026-01-13}
}

@inproceedings{vaswaniAttentionAllYou2017a,
  title = {Attention Is {{All}} You {{Need}}},
  booktitle = {Advances in {{Neural Information Processing Systems}}},
  author = {Vaswani, Ashish and Shazeer, Noam and Parmar, Niki and Uszkoreit, Jakob and Jones, Llion and Gomez, Aidan N and ukasz Kaiser, {\L} and Polosukhin, Illia},
  year = 2017,
  volume = {30},
  publisher = {Curran Associates, Inc.},
  doi = {10.48550/arXiv.1706.03762}
}

@article{vergeerMeasuringCalibrationLikelihoodratio2021,
  title = {Measuring Calibration of Likelihood-Ratio Systems: {{A}} Comparison of Four Metrics, Including a New Metric {{devPAV}}},
  shorttitle = {Measuring Calibration of Likelihood-Ratio Systems},
  author = {Vergeer, Peter and {van Schaik}, Yara and Sjerps, Marjan},
  year = 2021,
  month = apr,
  journal = {Forensic Science International},
  volume = {321},
  issn = {0379-0738},
  doi = {10.1016/j.forsciint.2021.110722},
  urldate = {2026-01-13}
}

@article{wilksExtendingLogisticRegression2009,
  title = {Extending Logistic Regression to Provide Full-Probability-Distribution {{MOS}} Forecasts},
  author = {Wilks, Daniel S.},
  year = 2009,
  journal = {Meteorological Applications},
  volume = {16},
  number = {3},
  pages = {361--368},
  issn = {1469-8080},
  doi = {10.1002/met.134},
  urldate = {2026-01-13},
  copyright = {Copyright \copyright{} 2009 Royal Meteorological Society},
  langid = {english}
}

@article{wrightUsingWordNgrams2017a,
  title = {Using Word N-Grams to Identify Authors and Idiolects: {{A}} Corpus Approach to a Forensic Linguistic Problem},
  shorttitle = {Using Word N-Grams to Identify Authors and Idiolects},
  author = {Wright, David},
  year = 2017,
  journal = {International Journal of Corpus Linguistics},
  volume = {22},
  number = {2},
  pages = {212--241},
  issn = {1384-6655, 1569-9811},
  doi = {10.1075/ijcl.22.2.03wri},
  urldate = {2026-01-22},
  langid = {english}
}

@article{niniGrammarBehavioralBiometric2026,
	title = {Grammar as a behavioral biometric: using cognitively motivated grammar models for authorship verification},
	author = {Nini, Andrea and Halvani, Oren and Graner, Lukas and Titze, Sophie and Gherardi, Valerio and Ishihara, Shunichi},
	year = {2026},
	month = {03},
	date = {2026-03-03},
	journal = {Humanities and Social Sciences Communications},
	pages = {455},
	volume = {13},
	number = {1},
	doi = {10.1057/s41599-025-06340-3},
	url = {https://www.nature.com/articles/s41599-025-06340-3},
	langid = {en}
}

@article{morrison2012,
	title = {Database selection for forensic voice comparison},
	author = {Morrison, Geoffrey and Ochoa, Felipe and Thiruvaran, Tharmarajah},
	year = {2012},
	month = {01},
	date = {2012-01-01},
	journal = {Proceedings of Odyssey 2012: The Language and Speaker Recognition Workshop}
}

@book{tukeyExploratoryDataAnalysis1977,
	title = {Exploratory data analysis},
	author = {Tukey, {John W. (John Wilder)}},
	year = {1977},
	date = {1977},
	publisher = {Addison-Wesley Pub. Co.},
	url = {http://archive.org/details/exploratorydataa0000tuke_7616},
	address = {Reading, Massachusetts},
	langid = {eng}
}

@inproceedings{halvaniAssessingApplicabilityAuthorship2019,
	title = {Assessing the Applicability of Authorship Verification Methods},
	author = {Halvani, Oren and Winter, Christian and Graner, Lukas},
	year = {2019},
	month = {08},
	date = {2019-08-26},
	publisher = {Association for Computing Machinery},
	pages = {1{\textendash}10},
	series = {ARES '19},
	doi = {10.1145/3339252.3340508},
	url = {https://dl.acm.org/doi/10.1145/3339252.3340508},
	address = {New York, NY, USA}
}

@inproceedings{halvaniAuthorshipVerificationAbsence2018,
	title = {Authorship Verification in the Absence of Explicit Features and Thresholds},
	author = {Halvani, Oren and Graner, Lukas and Vogel, Inna},
	year = {2018},
	month = {03},
	date = {2018-03-01},
	doi = {10.1007/978-3-319-76941-7_34},
	langid = {en}
}

@inproceedings{noeckerjrDistractorlessAuthorshipVerification2012,
	title = {LREC 2012},
	author = {Noecker Jr, John and Ryan, Michael},
	editor = {Calzolari, Nicoletta and Choukri, Khalid and Declerck, Thierry and {Do{\u{g}}an}, {Mehmet U{\u{g}}ur} and Maegaard, Bente and Mariani, Joseph and Moreno, Asuncion and Odijk, Jan and Piperidis, Stelios},
	year = {2012},
	month = {05},
	date = {2012-05},
	publisher = {European Language Resources Association (ELRA)},
	pages = {785{\textendash}789},
	url = {https://aclanthology.org/L12-1090/},
	address = {Istanbul, Turkey}
}

\pagebreak
\appendix
\section{LambdaG Algorithm}
\label{ap-a}
\begin{algorithm}[H]
\caption{LambdaG}
\begin{algorithmic}
\Statex \textbf{Input:} $Q$, $K$, $\mathbb{R}$, $N$, $r$
   \Statex 
\textcolor{gray}{$Q$ is the questioned text; $K$ is the known-author text; $\mathbb{R}$ is a set of reference texts ($\mathbb{R}$ = $R^{1}$, $R^{2}$, \dots); $N$ is the order of the model; $r$ is the number of repetitions} 
\Statex

\State \textbf{Function} $posnoise(D)$ 
\textcolor{gray}{Returns a POS-noise version of $D$, where content words are replaced by their part-of-speech tags}
         
\State \textbf{Function} $sent(D)$ 
\textcolor{gray}{Returns the set of all tokenized sentences in document(s) $D$}

\State \textbf{Function} $sample(\mathbb{S}, n)$
\textcolor{gray}{Randomly samples $n$ tokens from the set $\mathbb{S}$}

\State \textbf{Function} $KN(\mathbb{S}, N)$
\textcolor{gray}{Train an n-gram language model of order N with the set of sentences $\mathbb{S}$ using Kneser-Ney smoothing}

\Statex
\Statex\textcolor{gray}{Apply POSNoise to al respective documents and construct tokenized sentences}

\State $\mathbb{S}_Q \gets sent(posnoise(Q))$
\State $\mathbb{S}_K \gets sent(posnoise(K)) $
\State $\mathbb{S}_{R} \gets sent(posnoise(\mathbb{R}))$

\Statex
\Statex\textcolor{gray}{Build Grammar Model for the known author}
\State $G_K \gets KN(\mathbb{S}_K, N)$

\Statex
\Statex\textcolor{gray}{Build reference grammar models} 
\For{$i \gets 1$ to $r$} \do{}
  \State $\mathbb{S}_i \gets sample(\mathbb{S}_{R}, |\mathbb{S}_K|)$
  \State $G_{R}^{(i)} \gets KN(\mathbb{S}_i, N)$
\EndFor

\Statex \textcolor{gray}{Calculate the $\lambda_G$ (the log-likelihood ratio of the Grammar Model) over $\mathbb{S}_Q$} 
\Statex ${\lambda_G}(\mathbb{S}_Q \gets 0$ 
\For{$S_i \in$ $\mathbb{S}_Q$} \do{}
\textcolor{gray}{Decompose $S_i$ into a sequence of tokens}
\State $(t_1, t_2, \dots,t_z) \gets S_i$
\For{$j \gets 1$ to $z$} \do{}
\textcolor{gray}{Calculate mean $\lambda_G$ for $t_j$ over reference Grammar Models}
 \For{$i \gets 1$ to $r$} \do{}
 \State $\lambda_G(\mathbb{S}_Q) \gets \lambda_G(\mathbb{S}_Q) + \frac{1}{r}\log\frac{P(t_j|t_{<j}; G_K}{P(t_j|t_{<j}; G_j}$
\EndFor
\EndFor
\EndFor
\Statex \Return $\lambda_G(\mathbb{S}_Q)$

\end{algorithmic}
\end{algorithm}

\end{document}